\documentclass[a4paper,fleqn]{cas-dc}
\usepackage[sort,numbers]{natbib}

\usepackage{enumitem}

\usepackage{CJK}

\usepackage{ulem}
\usepackage{subfigure}

\usepackage{amsmath}
\usepackage{ulem} 

\usepackage{multicol}

\usepackage{tabularx}
\usepackage{xcolor}
\definecolor{Gray}{gray}{0.95}

\usepackage{pifont}

\def\tsc#1{\csdef{#1}{\textsc{\lowercase{#1}}\xspace}}
\tsc{WGM}
\tsc{QE}
\tsc{EP}
\tsc{PMS}
\tsc{BEC}
\tsc{DE}

\begin{document}
\let\WriteBookmarks\relax
\def\floatpagepagefraction{1}
\def\textpagefraction{.001}

\shortauthors{Q. Mao et~al.}

\title [mode = title]{Attend and Select: A Segment Selective Transformer for Microblog Hashtag Generation}

\author[1,2]{Qianren Mao}[orcid=0000-0003-0780-0628
]


\ead{maoqr@act.buaa.edu.cn}

\author[1,2]
{Xi Li}
\author[3]{Bang Liu}{}
\author[4]{Shu Guo}
\author[1,2]{Peng Hao}{}

\author[1,2]
{Jianxin Li}
\cormark[1]

\author[4]{Lihong Wang}{}


\address[1]{Beijing Advanced Innovation Center for Big Data and Brain Computing, Beihang University, Beijing {\rm 100191}, China.}
\address[2]{The State Key Laboratory of Software Development Environment, Beihang University, Beijing, {\rm 100191}, China}
\address[3]{RALI \& Mila, University of Montreal, UK}
\address[4]{CENCERT/CC, Beijing, {\rm 210016}, China.}

\cortext[cor1]{Corresponding author}

\begin{abstract}
Hashtag generation aims to generate short and informal topical tags from a microblog post, in which tokens or phrases form the hashtags.
These tokens or phrases may originate from primary fragmental textual pieces (e.g., segments) in the original text and are separated into different segments. 
However, conventional sequence-to-sequence generation methods are hard to filter out secondary information from different textual granularity and are not good at selecting crucial tokens. 
Thus, they are suboptimal in generating more condensed hashtags. 
In this work, we propose a modified Transformer-based generation model with adding a segments-selection procedure for the original encoding and decoding phases. 
The segments-selection phase is based on a novel Segments Selection Mechanism (SSM) to model different textual granularity on global text, local segments, and tokens, contributing to generate condensed hashtags. 
Specifically, it first attends to primary semantic segments and then transforms discontinuous segments from the source text into a sequence of hashtags by selecting crucial tokens. 
Extensive evaluations on the two datasets reveal our approach's superiority with significant improvements to the extraction and generation baselines. 
The code and datasets are available at \url{https://github.com/OpenSUM/HashtagGen}.
\end{abstract}



\maketitle
\section{Introduction}
The adoption of hashtags in major micro-blogging services, including Twitter, Facebook, and Sina Weibo, is strong evidence of its importance in facilitating information acquisition, communication, and diffusion. 
According to a study conducted by MarTech~\cite{bockler2014integrating}, 71\% of people on social media platforms use hashtags regularly, and almost half of those people explore related content by clicking on the provided hashtags. 
However, lots of cases in the massive microblog lack user-provided hashtags. 
For example, less than 15\% tweets contain at least one hashtag~\cite{WangWLZZ11,WangLKLS19}. 
Besides, the low-quality and irregular hashtags affect the social platform's acquisition and management of information.
There is an increasing demand for managing large-scale microblog contents, and tagging microblog is an effective means for information retrieval and content management. 
However, human annotation is time-consuming and costly. Besides, the  artificial construction may produce intentionally misleading tags, and the produced tags are inconsistent with the semantics that the post conveys.

Hashtag generation aims to summarize the main idea of a microblog post and generate short and informal topical tags.
Most previous works focus on keyphrases or keywords extraction methods~\cite{ChowdhuryCC20,GodinSNSW13,GongZH15,ZhangWGH16,ZhangLSZ18,LiangW0L21,GeroH21}.
However, extraction-based approaches fail to generate keyphrases that do not appear in the source document.  
These keyphrases are frequently produced by human annotators. 
Thus, hashtags may contain some new phrases or words which are not present in the posts. 
As shown in case A in Figure~\ref{hashtag}, the term `\textit{India}' in the hashtag does not exist in the posts.  
Keywords extraction methods may lose semantic coherence if there exists a slightly different sequence of keywords appearing in the posts. 
Another line of research is hashtags classification methods~\cite{GongZ16,HuangZGH16,WestonCA14,ZhangWHHG17} from the given tag-catalogs. 
These methods can not produce a hashtag that is not in the candidate catalogs list. 
In reality, a wide variety of hashtags can be created daily, making it impossible to cover all hashtags with a fixed candidate catalogs list.

There are some works~\cite{ChenZ0YL18,ChenGZKL19,MengZHHBC17,YeW18} that propose a  keyphrase generation task for regular textual documents and use the sequence-to-sequence framework for the generation. 
The recent works~\cite{WangLKLS19,ZhangLSZ18,WangLCKLS19} have transplanted the keyphrase generation task to the hashtag generation for Tweets.
Such methods suffer from long-term semantic disappearance existing in a recurrent neural network.
Hence, they are incapable of capturing  semantics among crucial tokens and not good at distilling critical information, either.

To show the textual features of post, we give two Twitter posts as shown in Figure~\ref{hashtag} and we partition them with 5 consecutive tokens.  Hashtags often appear in the tail of the post. 
Specifically, we have the following observations: 
\begin{itemize}[leftmargin=*]
\item Different keywords arranged in a hashtag may originate from various segments. 
More than 61.63\% of the words (excluding stop words) of hashtags appear in three or more different Chinese Weibo segments. 
Those segments highlighted in the dash line (which are coarse-grained) contain crucial tokens (which are fine-grained), reflecting the primary semantics of their hashtag.  
\vspace{-0.05in}
\item The critical tokens from different segments are usually discontinuous. 
In other words, there usually exist nested and hierarchical semantic dependencies among these crucial tokens. 
For example, in Case B, the `\textit{This new advancements}' refers to `\textit{5G}' in the first segment. 
 \vspace{-0.05in} 
\end{itemize}


 \begin{CJK}{UTF8}{gkai}
\begin{figure*}[width=2.08\linewidth,cols=4,pos=h]
  \centering
  \footnotesize
  \begin{tabular}{@{}p{17.45cm}@{}}
  \toprule
  \textbf{Case A, a Twitter post with its hashtags}：
  \dashuline{\textcolor{blue}{\textbf{Maharashtra}} also contributes to almost} 40\% of total deaths due \dashuline{to \textcolor{blue}{\textbf{coronavirus}} , with \textcolor{blue}{\textbf{Mumbai}}} contributing the most in Maharashtra. 78,761 cases in 24 hours , a new global record . \textcolor{blue}{\textbf{\#Coronavirus in India}}.      
  \\ 
  \bottomrule
  \textbf{Case B, a Twitter post with its hashtags}： 

  \dashuline{The \textcolor{blue}{\textbf{5G}} race is on} , but are carriers up to the challenge? This \dashuline{new advancements will \textcolor{blue}{\textbf{bring}} a} {} \dashuline{\textcolor{blue}{\textbf{wealth of new opportunities}} along} with their own security challenges .  \textcolor{blue}{\textbf{\#5G Bring New Value \#innovation}}                     
  \\ 
  \bottomrule
  \end{tabular}
  \caption{Illustrations of the two Twitter posts with their hashtags. The segment's length is fixed to be 5-tokens in the two cases. 
  Keywords or keyphrases (in bold blue) are distributed in several segments which are  highlighted with a dash line. 
  In the SNS platform, the post is always written with informative and short hashtags, which can be treated as natural annotation hashtags. }
  \label{hashtag}
\end{figure*}
\end{CJK}

To solve the issues mentioned above, we propose an end-to-end generative method \underline{Seg}ments Selective \underline{Tr}ansfor\underline{m}er (\textbf{S{\footnotesize EG}T{\footnotesize RM}}) to select crucial segments, and use these segments for hashtag generation. 
We introduce a novel \underline{s}egment attention based \underline{s}election \underline{m}echanism (\textbf{SSM}) to attend and select key contents.
In the segment selection mechanism, the input text is split into multiple blocks (segments) with a fixed length. 
The global document representation that represents the whole text, is customized by a special token prepended in front of the text. 
The local segmental representation that represents the whole segment, is customized by another special token inserted in front of each segment in the text. 
Then, the segment selection mechanism is calculated by a similarity score between the global textual representations and multiple local segmental representations. 
The top \(k\) similarity score allows the model to attend the top \(k\) sorted segments. And the model can select key segmental tokens as inputs to be feed into the decoder for a generation.











We propose two kinds of selection mechanisms (i.e., soft-based and hard-based) to select dominant textual representations which will be fed into the decoder. 
In the \textbf{soft-based SSM}, the selected targets are multiple segmental tokens of [SEG] with their collateral textual tokens. 
However, in the \textbf{hard-based SSM}, the selected targets will be multiple segmental [SEG] themselves without their collateral textual tokens.
Hard-based SSM is ultimately a hierarchical way (more hierarchical than soft-based SSM) to model the compositionality of segmental representations. 
Both mechanisms attend to segments and select tokens, in which the hierarchical segmental representations are used to capture complex or nested relations of segmental tokens in a hierarchical manner.
To summarize, our main contributions include:

\begin{itemize}[leftmargin=*]
 \item  We propose a segment selective Transformer (termed as \textbf{S{\footnotesize EG}T{\footnotesize RM}}) which extends the sequence-to-sequence model for the task of hashtag generation.
\vspace{-0.1in}
\item  We propose two selection mechanisms: a soft-based one and a hard-based one for segments selection. 
Two mechanisms are aware of filtering out secondary information by attending to primary segments and selecting crucial tokens from segments in a hierarchical manner.
\vspace{-0.1in} 
\item  Our model has achieved superior performance to the strong baselines on two newly constructed large-scale datasets. 
Notably, we obtain \textbf{significant improvements}  for Chinese Weibo and English Twitter hashtag generation. 
\end{itemize}

\section{Related Works for Hashtag Generation}
\subsection{Keyphrase Generation Methods}
It should be noted that there are also slight differences between the two tasks of hashtag generation and keyphrase generation.
The first one may summarize a short microblog in Social Networking Services (SNS), but the keyphrase generation task~\cite{Ahmad0LC20,CanoB19,ChenCLK20,ChenGZKL19,MengZHHBC17,SwaminathanZMGS20} is to generate phrases from a news document.

The keyphrase generation mainly generates multiple discontinuous words or phrases. 
In contrast, a hashtag can be a phrase-level short text that describes the main idea of a microblog post. Thus, these hashtags require a generation system that can rephrase or paraphrase keywords or keyphrases grammatically for the generation. 
But beyond that, hashtags could also be single keywords or keyphrases without any reorganization. 

The similarity between the two tasks is that they can generate keywords or keyphrases as the goal of the task. Since that, the hashtag generation task can be regarded as a branch of the keyphrase extraction task. As described by ~\citet{WangLKLS19}, hashtags can be used to reflect not only keyphrases but also topics~\cite{LiLGHW16,YanGLC13}.

\subsection{Hashtag Generation Methods}
Most early works in hashtag generation focus on tag selection for the post from pre-defined tag  candidates~\cite{GongZ16,HuangZGH16,JavariHHRC20,ZhangWHHG17}. 
However, hashtags usually appear in neither the target posts nor the given candidate lists. 
~\citet{WangLCKLS19} and ~\citet{WangLKLS19}  are the first to approach hashtag generation with a generation framework. 
In doing so, phrase-level hashtags beyond the target posts or the given candidates can be created.  
~\citet{WangLKLS19} realize hashtag generation by a topic-aware generation model that leverages latent topics to enhance valuable features.
~\citet{ZhangLSZ18} and ~\citet{WangLCKLS19} propose to jointly model the target posts and the conversation contexts with bidirectional attention. 

However, their works require massive external conversion snippets or relevant tweets for modeling. The generated results are directly affected by noisy conversations or other tweets. 
In reality, the external text does not necessarily exist, and there is a high cost for external texts annotation.
In addition, the dataset~\footnote{Their datasets will be introduced in section~\ref{DataConstruction} in detail.} they released also have disadvantages, such as small-scale, sparse distribution, and insufficient related domain.

\begin{figure*}[width=2.1\linewidth,cols=4,pos=h]
  \centering
  \includegraphics[scale=0.495]{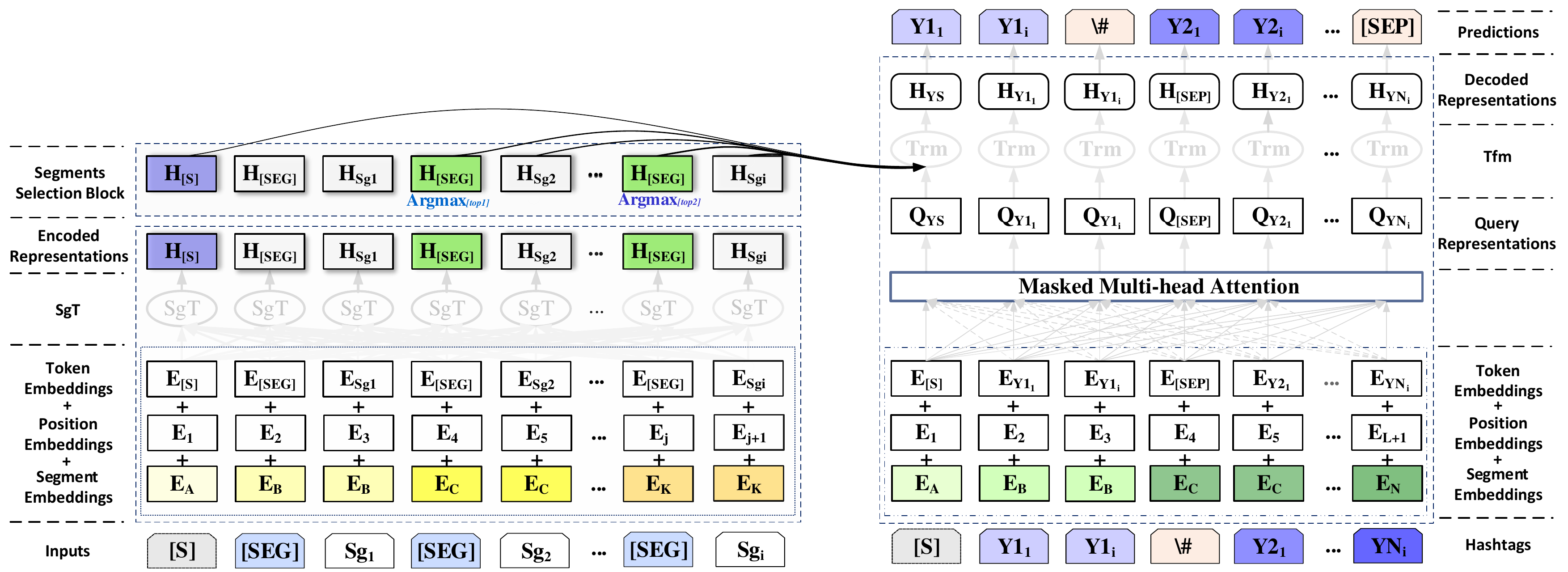}
  \caption{The overview architecture of the Segments-selection based Transformer  (\textbf{S{\footnotesize EG}T{\footnotesize RM}}) augmented by a segments-selection block for hashtag generation. For simplification, we use  
  ${\rm Sg}_{i}$ to represent a bunch of tokens ${\rm Sg}_{i}=[token^{i}_{1}, token^{i}_{2}, ...]$  in the \(i\)-th segment.
  Target hashtags are separated by `\(\setminus\)\#'.} 
  \vspace{-0.05in}
  \label{fig:model}
\end{figure*}

\section{Methodology}
\noindent\textbf{Problem Formulation}. 
 We define the microblog hashtag generation as a given microblog post, automatically generating a sequence consisting of condensed topic hashtags. 
 Each hashtag is separated by separators `\(\setminus\#\)'.  
The task can be regarded as one of the sub-classes of text generation. 
However, the hashtag generation system is applied to learning the mapping from a post to multiple target hashtags.

\noindent\textbf{Segmentation}. 
Segments are textual blocks splited by position with fixed length for input sentence. 
To ensure efficiency for batch processing and dimension alignment, we set each segment's length being fixed \footnote{Fixed length segmentation can guarantee the high-efficiency of batch processing and dimension alignment. As for indefinite length segmentation with consideration for syntax, such as clause-based segmentation, we leave it to future work.}. 
To represent an individual segment, we insert an external [SEG] token at the start of this segment ($[{\rm SEG}]$, ${\rm Sg}_{i}$). 
The  ${\rm Sg}_{i}=[token1,...,tokenk]$ refers to a segment with its tokens and \(k\) is the length of the segment.
The [SEG] follows a short sequential  collateral textual tokens and is used to aggregate segmental semantic representations.

\noindent\textbf{Model Architecture}. 
As shown in Figure~\ref{fig:model}, our S{\footnotesize EG}T{\footnotesize RM} consists of three phases: encoder (bottom left dotted box), segments-selector (top left dotted box), and hashtag generator (right dotted box). 
In the encoder, we prepend a token [S]\footnote{It is similar to the usage of [CLS] in the pre-trained language models, that is, to represent the global semantics of the text. However, [S] is trained from scratch.} in front of the text and use it to obtain global textual representations.  
The segments-selector (which will be introduced in detail in section~\ref{SSM}) selects multiple segments and recombines them into a new sequence as the decoder's input to generate hashtags in an end-to-end way.  
To simultaneously predict multiple hashtags and determine the suitable number of hashtags in the generator, we follow the settings~\cite{YuanWMTBHT20} by adopting a sequential decoding method to generate one sequence consisting of multiple targets and separators. 
We insert multiple `\(\setminus\)\#' tokens as separators. 
During generation, the decoder stops predicting when it encounters the terminator [SEP].

\subsection{Encoder with Segmental Tokens}
\label{EST}
The segmentation to represent different granular text has been successfully used in language models (LMs), such as BERT~\cite{devlin2018bert,abs-1906-04341}). 
However, segment embedding of LMs is used to distinguish different sentences in natural language inference tasks. 
Unlike segmentation in BERT, we aim to represent different segmental sequences by inserting multiple special tokens [SEG].  
We assign interval segment embedding [${\rm E}_{A}$, ${\rm E}_{B}$, ${\rm E}_{C}$, ..., ${\rm E}_{K}$] to differentiate multiple segments. The visualization of this construction can be seen in Figure~\ref{fig:model}.
Each token's embedding is the sum of initial token embedding, position embedding, and segment embedding. 
The input ${\rm I}_{X}$ of a post text is represented as:
\begin{equation}
{\rm I}_{X}=\! \mathcal{I}\left \{ [{\rm S}],[{\rm SEG}],{\rm Sg}_{1},...,[{\rm SEG}],{\rm Sg}_{i},..., \right \},
\end{equation}
where $\mathcal{I}\left \{ \cdot  \right \}$ is an \textit{insert} process with inserting [S] and [SEG] into text ${\rm X}$.

\begin{figure*}
  \begin{minipage}[t]{.48\linewidth}   
  \centering
      \subfigure[Text attention]{
  \includegraphics[width=2.7in]{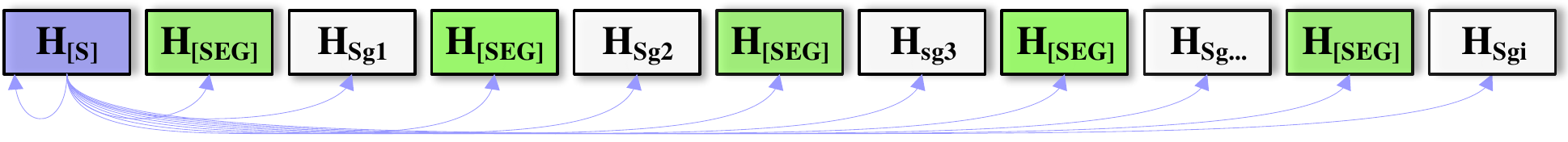}}
      \label{fig:side:a}  
  \end{minipage}
  \begin{minipage}[t]{.48\linewidth}   
      \centering   
          \subfigure[Segment attention]{
  \includegraphics[width=2.7in]{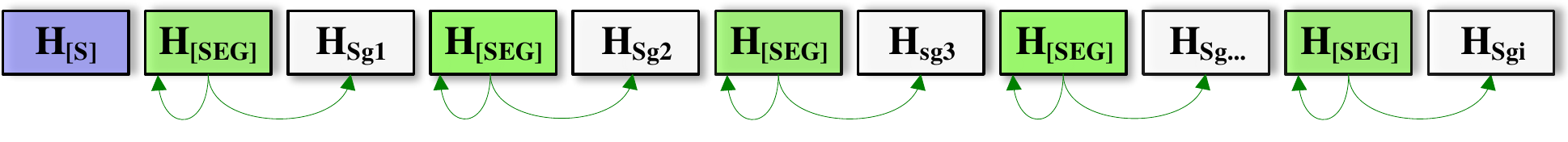}}
      \label{fig:side:b} 
  \end{minipage}

  \begin{minipage}[t]{1\linewidth}   
    \centering   
        \subfigure[Token attention]{
\includegraphics[width=2.7in]{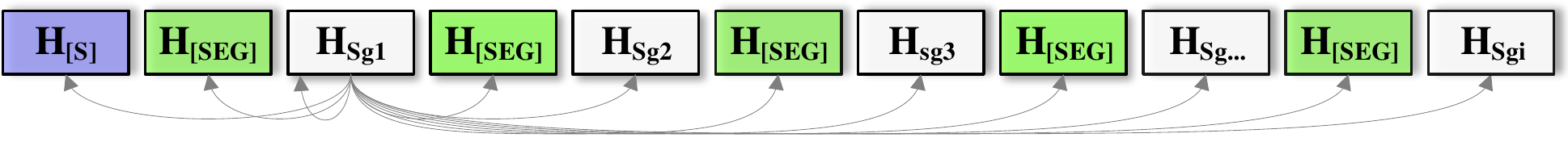}}
    \label{fig:side:b}    
\end{minipage}
    \caption{An illustration of `Attend' operation with attention mechanisms to different textual granularity: text, segments and tokens in SgT.}
    \label{similarityattention}
\end{figure*}

\begin{figure*}
\centering
\begin{minipage}[t]{.45\linewidth}  
\centering
    \subfigure[Soft segment selection]{
\includegraphics[width=2.7in]{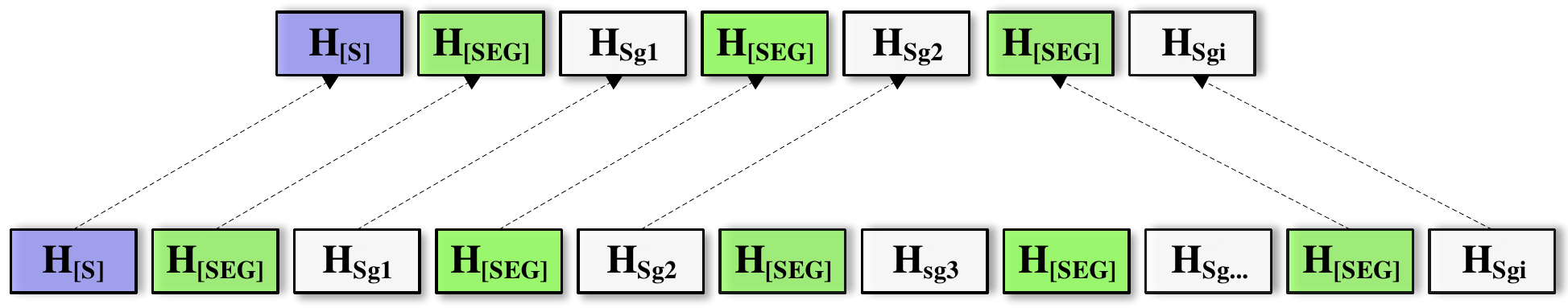}}
\end{minipage}
\begin{minipage}[t]{.45\linewidth}   
    \centering   
        \subfigure[Hard segment selection]{
\includegraphics[width=2.7in]{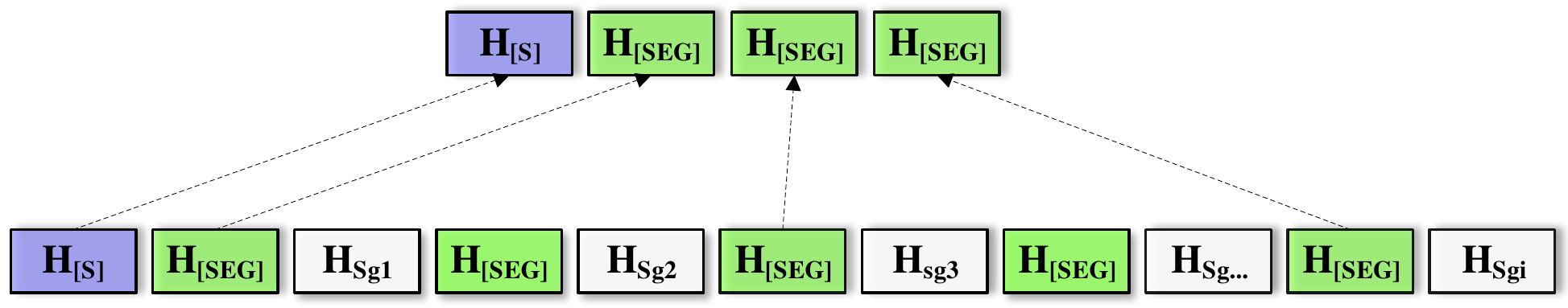}}
\end{minipage}
   \caption{An illustration of `Select' operation of the soft and hard ways in segment selection block. The selected targets are multiple segmental tokens of [SEG] with their collateral textual tokens in soft-based SSM. However, in the hard-based SSM, the selected targets are  multiple [SEG] themselves without any  collateral textual tokens.}
  \label{similarity}
  \vspace{-0.1in}
\end{figure*}

Our S{\footnotesize EG}T{\footnotesize RM}'s encoder (termed as \textbf{SgT}) is equipped with three kinds of attention mechanisms to different textual granularity: text, segments, and tokens, as shown in Figure~\ref{similarityattention}.

In SgT, to let [SEG] learn the local semantical representations, we assign a mask vector to each token based on the fixed segment length. 
Taking Figure~\ref{similarityattention} (b) as an example, the mask vector of the first segment is [0, 1, 1, 0, 0, 0, 0, 0, 0, 0, 0]. 
After obtaining the mask vector for each token in a specific segment, we stack them to form a \(n*n\) mask matrix \({\rm M}_{Sg_{i}}\) and calculate local \(i\)-th segment attention with the equation below. 
For simplification, we write it in the one-head form. 
\begin{equation}
 Attention_{Sg_{i}}\left ({\rm Q,K,V} \right )=softmax\left ( \frac{{\rm QK}^{\top}{\rm M}_{Sg_{i}}}{\sqrt{d_{k}}} \right ){\rm V},
\end{equation}
where \({\rm Q}\) refers to the queries, \({\rm K}\) the keys, \({\rm V}\) the values. \(\frac{1}{\sqrt{d_{k}}}\) is a scaling factor and \(d_{k}\) is dimension of queries and keys.      
The text attention and token attention are the same as the multi-head self-attention in vanilla Transformer.

In the encoder SgT, textual representations are learned hierarchically.
In other words, the model is aware of the hierarchical structure among different textual granularity. 
Lower SgT layers represent adjacent segments, while higher layers obtain contextual multi-segments representations.

\subsection{Segments Selection Mechanism}
\label{SSM}
The upper layer of the encoder is a segments-selection block used to select critical tokens. 
In the block, by calculating the semantic similarity weight, (the greater the similarity, the closer the semantic space distance) between ${\rm H}_{[S]}$ and  ${\rm H}^{i}_{[SEG]}$, important segmental pieces are selected. 
The vector ${\rm H}^{i}_{SEG}$, is used as the local representations for $[{\rm Sg}_{i}]$. 
Similarly, the global representations for the post text are \normalsize{${\rm H}_{[S]}$}. 

Thus, given a pair of ${\rm H}_{[S]}$ and ${\rm H}^{i}_{[SEG]}$, the similarity weight is calculated by \normalsize{$S_{i}=f(\rm H_{[S]},\rm H^{i}_{[SEG]})$}.  
There are multiple similarity functions \normalsize{$f$} in our model. 
We introduce four common functions that can serve as \normalsize{$f$}:
\begin{itemize}[leftmargin=*]
\item {Euclidean-distance Similarity (\#ES)} 
\item{Cosine-distance Similarity (\#CS)} 
\item {Mahalanobis-distance Similarity (\#MasS)}
\item {Manhattan-distance Similarity (\#MhtS)} 
\end{itemize}
After calculating the similarity score by \normalsize{$f$}, we can select top \(k\) segments ${\rm H}^{i}_{[SEG]}$ with the highest similarity score. 
Then, segmental tokens are collected to form a new sequence of hidden representations \({\rm H}_{X^{s}}\).

As the two aforementioned selection methods to model the segmental compositionality, the soft segment selection method will select top \(k\) [SEG] and their collateral textual tokens. 
As Fig~\ref{similarity}.(a) shown, the selected sequence is \({\rm H}_{X^{s}}=[{\rm H}_{S},{\rm H}_{[SEG]},{\rm H}_{\rm sg1},...,{\rm H}_{[SEG]},{\rm H}_{\rm sgi}]\), where the $\left [ \cdot  \right ]$ is an appending operation. 
The \({\rm H}_{\rm sgi}\) refers to a sequence of tokens' hidden representations $\left[{\rm H}^{i}_{x1},{\rm H}^{i}_{x2},..,{\rm H}^{i}_{xn} \right]$. 
The hard segment selection method will only select top \(k\) [SEG]. 
As the Fig~\ref{similarity}.(b) shown, the selected sequence is \({\rm H}_{X^{s}}=[{\rm H}_{S},{\rm H}_{[SEG]},{\rm H}_{[SEG]}]\).
The hard-based SSM does not select any original word as the decoder's input and is to model the segmental compositionality. 
Further, these selected pieces are fed into the decoder for hashtag generation.

\subsection{Decoder for Hashtag Generation}
\label{DHG}
Inspired by the sequential decoding method~\cite{YuanWMTBHT20}, we insert separators `\#' in the middle of each target and insert the [SEP] at the end of the sequence.
In doing so, the method can simultaneously predict hashtags and determine a suitable number of hashtags.
The multiple hashtags are obtained after splitting the sequence by separators.
Thus, hashtags is formulated by \normalsize{$[{\rm Y1},{\rm \setminus\#},{\rm Y2},{\rm \setminus\#},...,{\rm YN},{\rm [SEP]}]$}, where \({\rm Yi}\) is a hashtag sequence composed of \(x\) tokens: \({\rm Y1}_{1},{\rm Y1}_{i},...,{\rm Y1}_{x}\). 
\(\setminus\)\# is the separator and [SEP] is the terminator. 

We use a standard Transformer decoder for hashtag generation. 
Specifically, the input of the decoder, as shown in the Figure~\ref{fig:model}, is \({\rm I}_{Y}=\left \{ [{\rm S}],{\rm Y1}_{1},{\rm Y1}_{i},\setminus\#...,{\rm YN}_{i},...,[SEP]\right\}\).
The token [S] aggregating global representations of source posts is used as a `begin-of-hashtag' vector. 
The Transformer decoder transforms the ${\rm I}_{Y}$ and encoder's hidden states ${\rm H}_{X}$ into its representations ${\rm H}_{Y}$. 
${\rm H}_{Y}$ is then processed by the softmax operation for token prediction.

\begin{table}[tb]
  \centering
  \scriptsize
  \caption{Data statistics for Weibo hashtag generation. AvgSourceLen denotes the average length of all posts, and AvgTargetLen is the average length of hashtags. CovSourceLen (95\%) means that more than 95\% of the sentences have the exact length.
  }
  \setlength{\tabcolsep}{0.9mm}{
    \begin{tabular}{@{}l|lll|llr@{}}
  \toprule
  \multirow{2}{*}{DATASET} & \multicolumn{3}{c|}{
  Weibo:\textbf{WHG} Dataset}  & \multicolumn{3}{c}{Twitter:\textbf{THG} Dataset} \\
  \cline{2-7}
  & \textbf{Train}  & \textbf{Dev.}  & \textbf{Test} 
  & \textbf{Train}  & \textbf{Dev.}  & \textbf{Test}
  \\ 
  \cline{1-7}
  Post-hashtag pairs 
  &312,762 &2,000 &2,000  
  & 204,039 & 11,335  & 11,336                                   \\ 
  \hline
  CovSourceLen(95\%)    & 141       & 137     & 145  &  46       & 47      & 46   \\   
   AvgSourceLen      & 75.1      & 75.3   &  75.6  
   & 23.5      & 23.8  &  23.5 \\
  CovTargetLen(95\%)    & 8       & 8      & 8   & 31       & 30    & 31   \\  
  AvgTargetLen    & 4.2       & 4.2      & 4.2   & 10.1     & 10.0     & 10.0  \\
  \cline{1-7}
  \multicolumn{1}{@{}l|}{AvgHashtags}     
  &   1     & 1   &  1   &   4.1     & 4.1   &  4.1  \\  
  \bottomrule
  \end{tabular}}
  \label{tab:tableWHG}
\end{table}

\section{Experiment Setup}
\subsection{Implementation Details}
\label{implementdetails}
We implement 12 deep layers in both the encoder and decoder. 
The embedding size and hidden size of the encoder and decoder are set to 768.
The number of self-attention heads is 12.
We use the cross-entropy loss to train the models. 
The optimizer is Adam with a learning rate  1e-4, L2 weight decay $\beta1$ = 0.9, $\beta2$ = 0.999, and $\epsilon$ = 1e-6. 
The dropout probability is set to 0.1 in all layers. 
Following OpenAI GPT and BERT, we use a $gelu$ activation which performs better than the standard $relu$. 
The gradient clipping is applied with range [-1, 1] in the encoder and decoder. 
We implement a linear warmup with a Ratio~\cite{howard2018universal} of 32. 
The Ratio specifies how much smaller the lowest learning rate compares with the maximum one.
The proportion of warmup steps is 0.04 on the WHG and THG datasets. 
We use LTP Tokenizer~\footnote{\url{https://github.com/HIT-SCIR/ltp}} and RoBERTa's FullTokenizer~\cite{devlin2018bert} for preprocessing Chinese Weibo characters and English Twitter words, respectively. 
The length of the input and output is considered to be CovSourceLen and CovTargetLen in Table~\ref{tab:tableWHG}.

All models are trained on 4 GPUs\footnote{Tesla V100-PCIE-32GB}. 
We select the best 3 checkpoints on the validation set and report the average results on the test set. 
The hyperparameters of the number of Top \(k\) selected segments will be introduced in the experimental analysis (section~\ref{CDS}).


\subsection{Constructed Datasets}
\label{DataConstruction}
In previous hashtag generation works implemented by ~\citet{WangLCKLS19,WangLKLS19}, the annotation data is too small for the large Transformer, and has some deficiencies making it not satisfy the real-world situations and practical needs.
The existing Twitter dataset~\cite{WangLCKLS19} is built based on the TREC 2011 microblog track~\footnote{\url{https://trec.nist.gov/data/tweets/}}, and most of the tweets obtained by this tool are invalid now.  
The existed Chinese dataset~\cite{WangLCKLS19} also has obvious shortcomings. 
Firstly, the dataset contains only 40,000 posts for hashtag generation, and there is a long-tail distribution that 19.74\% of the hashtags are composed of one word.
Secondly, the lengths of 16.88\% posts are less than 10 words, and 84.90\% posts' lengths are less than 60, which is not consistent with Weibo's real-world data. 
Thirdly, in terms of content, entertainment-related text accounts for the majority of the dataset.
It is hardly practical through fine-tuning (whatever based on language models or based on our S{\scriptsize EG}T{\scriptsize RM}). 
Thus, there is an apparent semantic bias between the pre-training data (multiple domains) and their corpus (entertainment domain).

We construct two new large-scale datasets: the English Twitter hashtag generation (THG) dataset and the Chinese Weibo
hashtag generation (WHG) dataset.

The construction details are introduced in Appendix~\ref{DC}. 
We use 312,762 post-hashtag pairs for the training on WHG and
204,039 for the training on THG. 
As Table 1 shown, the average
number of hashtags is 1 in WHG and 4.1 in THG, and their total
sequence length is about 4.2 and 10.1, respectively. 
The lengths of Twitter hashtags are shorter than those of Weibo's hashtags. 
It should be noted that hashtags in the middle of a post are not considered as they generally act as semantic elements rather than topic words~\cite{ZhangWGH16,ZhangLSZ18,WangLKLS19}.

\subsection{Comparative Baselines}
\vspace{-0.02in}
The baselines can be classified into three types: keywords extraction method, neural selective encoding generator, and Transformer-based generator.
It should be noted that the existing two approaches~\citet{WangLCKLS19} and ~\citet{WangLKLS19} can not be directly used in our baselines.
The model proposed by ~\citet{WangLCKLS19}, can not be used directly for our data because the model needs external inputs (relevant tweets) to construct a neural topic module.
The model proposed by ~\citet{WangLKLS19}, needs external conversational contexts and has a conversation encoder. 

Two keywords extraction methods, unsupervised extraction or supervised generation are: 
\begin{itemize}[leftmargin=*]
  \item 
\textbf{Ext.TFIDF}~\footnote{\url{https://github.com/scikit-learn/scikit-learn}} is an classic extraction method. We extract 3 keywords for Weibo hashtag organization (2 keywords for organizing Twitter hashtag).
The number of words used to form a hashtag is in accordance with the average number of tokens in hashtags. 
Following the tokens' order in the text, the extracted keywords are organized.
\vspace{-0.05in}
\item 
\textbf{ExHiRD}~\cite{ChenCLK20} which is augmented with a GRU-based hierarchical decoding framework. 
Selective encoding models are neural generation methods based on the selection of key information pieces. 
\end{itemize}

We introduce two neural sequence-to-sequence generation methods which are kind of content selectors: 
\begin{itemize}[leftmargin=*]
\item 
\textbf{S{\small EASS}}~\cite{zhou2017selective} is based on selective attention mechanism, and is a LSTM-based sequence-to-sequence framework.  It consists of a sentence encoder, a selective gate network, and an attention equipped decoder. 
\vspace{-0.05in}
\item 
\textbf{B{\footnotesize OTTOM}U{\footnotesize P}}\footnote{We implement a state-of-the-art variant called `B{\scriptsize OTTOM}U{\scriptsize P} with D{\scriptsize IFF}M{\scriptsize ASK}' as B{\scriptsize OTTOM}U{\scriptsize P}.}~\cite{GehrmannDR18} is a Transformer-based model augmented with bottom-up selective attention. 
B{\scriptsize OTTOM}U{\scriptsize P} determines which phrases in the source document should be selected and then applies a copy mechanism only to the preselected phrases during decoding.
\end{itemize}

Another salient sequence-to-sequence generation baselines are the vanilla Transformer~\cite{VaswaniSPUJGKP17} and our two base models without any SSM.

\begin{itemize}[leftmargin=*]
\item  
\textbf{T{\footnotesize RANS}A{\footnotesize BS}} is trained with the same settings as in ~\citet{VaswaniSPUJGKP17}.
The vanilla Transformer is different from our two base models on parameter settings in that we abandon the two regularizations of every bias regularization and LayerNorm regularization, as BERT~\cite{abs-1906-04341} does. 
We choose to use BERT's activation function \(f_{c}\) = Gelu instead of Relu of the vanilla Transformer. 
\vspace{-0.05in}
\item 
\textbf{S{\footnotesize EG}T{\footnotesize RM} Softbase} and \textbf{S{\footnotesize EG}T{\footnotesize RM} Hardbase} are two base models which are used to investigate the effects of two kinds of SSMs.
For S{\footnotesize EG}T{\footnotesize RM} (soft), its based model is S{\footnotesize EG}T{\footnotesize RM} Softbase, which is S{\footnotesize EG}T{\footnotesize RM} without (w/o) any SSM. 
S{\footnotesize EG}T{\footnotesize RM} Softbase can be used as the ablation model.
For S{\footnotesize EG}T{\footnotesize RM} (hard), the base model is S{\footnotesize EG}T{\footnotesize RM} Hardbase. 
It selects all representations of segmental [SEG] and then feeds all [SEG] into the decoder. 
\end{itemize}

\begin{table*}[width=2.1\linewidth,cols=4,pos=h]
  \centering
  \caption{R{\footnotesize OUGE} F1 results of models on the Weibo and Twitter hashtag generation datasets. \!The R{\footnotesize OUGE} results are $means \!\pm\! S.D.\left ( n\!=\!3\right )$. The F@\(k\) is the result of the model with the highest R{\footnotesize OUGE} score.
  }
  \renewcommand\arraystretch{1.}
  \setlength{\tabcolsep}{2mm}{
    \begin{tabular}{@{}l|lllll|llllr@{}}
  \toprule
  \multirow{2}{*}{Models} & \multicolumn{5}{c|}{
  Weibo:\textbf{WHG} Dataset}  & \multicolumn{5}{c}{Twitter:\textbf{THG} Dataset} \\
  \cline{2-11}
  & \textbf{R{\scriptsize OUGE}-1} 
  & \textbf{R{\scriptsize OUGE}-2} 
  & \textbf{R{\scriptsize OUGE}-L} 
  & F1@1 & F1@5 
  &\textbf{R{\scriptsize OUGE}-1} 
  & \textbf{R{\scriptsize OUGE}-2} 
  & \textbf{R{\scriptsize OUGE}-L}  
  & F1@1 & F1@5 
  \\ 
  \cline{1-11}
  Ext.TFIDF            
  & 18.02 & 2.30 & 15.45  & 17.12   & 18.10
  & 12.47 & 1.21 & 12.47    & 12.45    & 16.00
  \\
  S{\scriptsize EASS}          
  & 28.19$\pm${\scriptsize.13} & 18.40$\pm${\scriptsize.31} & 27.87$\pm${\scriptsize.13}     & 20.07    & 21.00
  & 28.33$\pm${\scriptsize.32} & 18.77$\pm${\scriptsize.31} & 28.49$\pm${\scriptsize.61}  & 18.33    & 19.11\\
  ExHiRD  
  & 30.19$\pm${\scriptsize.12} & 19.40$\pm${\scriptsize.21} & 29.87$\pm${\scriptsize.12}    
  & 23.32    & 24.11
  & 29.17$\pm${\scriptsize.55} & 19.22$\pm${\scriptsize.71} & 28.54$\pm${\scriptsize.41} 
  & 20.52    & 22.41\\
  B{\scriptsize OTTOM}U{\scriptsize P}    
  & 34.33$\pm${\scriptsize.21} & 24.37$\pm${\scriptsize.17} & 35.14$\pm${\scriptsize.31}  & 26.32    & 25.32     
  & 42.29$\pm${\scriptsize.91} & 26.90$\pm${\scriptsize1.01} & 37.77$\pm${\scriptsize.59}
  & 22.41    & 22.77\\
  T{\scriptsize RANS}A{\scriptsize BS}        
  & 52.13$\pm${\scriptsize.11} & 46.62$\pm${\scriptsize.12} & 51.05$\pm${\scriptsize.11}    & 25.47    & 28.32 
  & 43.71$\pm${\scriptsize.17} & 27.18$\pm${\scriptsize.15} & 39.29$\pm${\scriptsize.14}  & 23.22    & 23.27 \\ 
  \hline 
   S{\scriptsize EG}T{\scriptsize RM} {\scriptsize Hardbase}                   
  & 54.53$\pm${\scriptsize.20} & 50.26$\pm${\scriptsize.13} &53.29$\pm${\scriptsize.22}    
  & 30.11    & 31.01    
  & 46.37$\pm${\scriptsize.11} & 30.71$\pm${\scriptsize.13} &41.73$\pm${\scriptsize.09}
  & 24.35    & 26.32\\ 
  S{\scriptsize EG}T{\scriptsize RM} {\scriptsize Hard}                 
  &55.40$\pm${\scriptsize.13} &51.32$\pm${\scriptsize.11} &54.12$\pm${\scriptsize.13}    
  & 30.73    & 31.76  
  & 47.25$\pm${\scriptsize.27} & 31.78$\pm${\scriptsize.33} &42.63$\pm${\scriptsize.21}
  & 25.31    & 27.07\\ 
   S{\scriptsize EG}T{\scriptsize RM} {\scriptsize Softbase}                   
  & 52.62$\pm${\scriptsize.18} & 48.70$\pm${\scriptsize.10} &51.41$\pm${\scriptsize.13}   
  & 28.62    & 30.58 
  & 50.00$\pm${\scriptsize.29} & 35.48$\pm${\scriptsize.19} &45.82$\pm${\scriptsize.17}
  & 26.02    & 28.59\\ 
  S{\scriptsize EG}T{\scriptsize RM} {\scriptsize Soft}                 
  &\textbf{55.51$\pm${\scriptsize.17}} &\textbf{51.28$\pm${\scriptsize.09}} &\textbf{54.30$\pm${\scriptsize.10}}   & \textbf{30.72}    & \textbf{32.21}     
  & \textbf{51.18$\pm${\scriptsize.19}} & \textbf{37.15$\pm${\scriptsize.12}} &\textbf{47.05$\pm${\scriptsize.31}}  & \textbf{27.17}    & \textbf{29.02} \\ 
  \bottomrule
  \end{tabular}}
\label{mainresults}
\end{table*}

\subsection{Evaluation Metric}
We use the official R{\footnotesize OUGE} script\footnote{\url{https://pypi.org/project/pyrouge/}} (version 0.3.1) as our evaluation metric.
We report R{\footnotesize OUGE} F1 to measure the overlapping degree between the generated sequence of hashtags and the reference sequence, including unigram, bigram, and longest common subsequence.
The reasons why we choose the R{\footnotesize OUGE} as our evaluation can be concluded as two aspects.
Firstly, the task aims to generate sequential hashtags, and R{\footnotesize OUGE} is a prevailing evaluation metric for the generation task. 
Secondly, we find multiple hashtags are helpful to reflect the relevance of the target post to the hashtag. 
For instance, although these hashtags (e.g., `\#farmers', `\#market', and `\#organic farmers') are not the same as the reference one (e.g., `\#organic farmers market'), they are usable.
The n-gram overlaps of R{\footnotesize OUGE} will not miss the highly available hashtags, but F1@K will since it only can evaluate the hashtags identical to the reference one.
To ascertain which correct tokens are not identical to the reference tokens but are copied from the source text, we test the n-gram overlaps between the generated text and the source text. 
This evaluation metric can identify the extraction ability of models. 

We also use F1@\(k\) evaluation (a popular information retrieval evaluation metric) to verify the ability of our model to normalize a single hashtag. 
F1@\(k\) compares all the predicted hashtags with ground-truth hashtags. 
Beam search is utilized for inference, and the top \(k\) hashtag sequences are leveraged to produce the final hashtags.
Here we use a beam size of 20, and \(k\) as 10.
Since our model can generate multiple hashtags (separated by \#) for a document, the final F1@\(k\) are tested for multiple hashtags.

\begin{table*}[width=2.1\linewidth,cols=4,pos=h]
  \centering
  \caption{R{\footnotesize OUGE} F1 results of models with different similarity metrics of SSM on the Weibo and Twitter hashtag generation datasets.}
  \renewcommand\arraystretch{1}
  \setlength{\tabcolsep}{5.4mm}{
    \begin{tabular}{@{}ll|lll|lllr@{}}
    \toprule
    \multicolumn{2}{l|}{\multirow{2}{*}{SSMs}} & \multicolumn{3}{c|}{
    Weibo:\textbf{WHG} Dataset}  & \multicolumn{3}{c}{Twitter:\textbf{THG} Dataset} \\
    \cline{3-8}
    \multicolumn{2}{l|}{} & \textbf{R{\scriptsize OUGE}-1} & \textbf{R{\scriptsize OUGE}-2} & \textbf{R{\scriptsize OUGE}-L} &
    \textbf{R{\scriptsize OUGE}-1} & \textbf{R{\scriptsize OUGE}-2} & \textbf{R{\scriptsize OUGE}-L}
    \\ 
    \cline{2-7}
     \hline
  \multirow{4}{*}{\rotatebox{90}{Hard}}
  &\#ES   & 54.34$\pm${\scriptsize.11} &49.89$\pm${\scriptsize.24}  & 53.17$\pm${\scriptsize.11}  
  & 46.03$\pm${\scriptsize.07} & 30.37$\pm${\scriptsize.10} &41.32$\pm${\scriptsize.13} 
  \\
  &\#CS                 & 54.91$\pm${\scriptsize.28}  &50.80$\pm${\scriptsize.23}  & 53.74$\pm${\scriptsize.36}  & 46.62$\pm${\scriptsize.21} & 31.12$\pm${\scriptsize.14} &42.20$\pm${\scriptsize.12}
  \\                    
  &\#MasS             & 54.53$\pm${\scriptsize.20}  &50.26$\pm${\scriptsize.15}  & 53.28$\pm${\scriptsize.17}  
  & 46.56$\pm${\scriptsize.13} & 30.89$\pm${\scriptsize.19} &41.85$\pm${\scriptsize.20}
  \\   
  &\#MhtS              & \textbf{55.40$\pm${\scriptsize.13}} &\textbf{51.32$\pm${\scriptsize.11}} &\textbf{54.12$\pm${\scriptsize.13}}  & \textbf{47.25$\pm${\scriptsize.27}} & \textbf{31.78$\pm${\scriptsize.33}} &\textbf{42.63$\pm${\scriptsize.21}}
  \\    
  \cline{1-8}
  \multirow{4}{*}{\rotatebox{90}{Soft}}
  &\#ES   & 44.74$\pm${\scriptsize.11} &40.89$\pm${\scriptsize.09}  & 43.67$\pm${\scriptsize.09}  
  &50.14$\pm${\scriptsize.09} &35.81$\pm${\scriptsize.12}  & 45.91$\pm${\scriptsize.11} 
  \\
  &\#CS                 & 53.83$\pm${\scriptsize.08}  &49.70$\pm${\scriptsize.13}  & 52.60$\pm${\scriptsize.06}  
  & 50.19$\pm${\scriptsize.13}  &35.76$\pm${\scriptsize.09}  
  & 46.00$\pm${\scriptsize.10} 
  \\
  &\#MasS             & 54.09$\pm${\scriptsize.05}  &50.05$\pm${\scriptsize.05}  & 52.92$\pm${\scriptsize.07}  
  & \textbf{51.18$\pm${\scriptsize.19}} & \textbf{37.15$\pm${\scriptsize.12}} &\textbf{47.05$\pm${\scriptsize.31}} 
  \\
  &\#MhtS              & \textbf{55.51$\pm${\scriptsize.17}} &\textbf{51.28$\pm${\scriptsize.09}} &\textbf{54.30$\pm${\scriptsize.10}} 
  & 50.86$\pm${\scriptsize.10}  &36.70$\pm${\scriptsize.11}  & 46.75$\pm${\scriptsize.09}\\    
  \bottomrule
  \end{tabular}}
  \label{selectionresults}
 \end{table*}

\begin{figure*}[width=2.1\linewidth,cols=4,pos=t] 
  \begin{minipage}[t]{0.245\linewidth}   
      \centering   
      \subfigure[Soft SSM on WHG]{
  \includegraphics[width=1.4in]{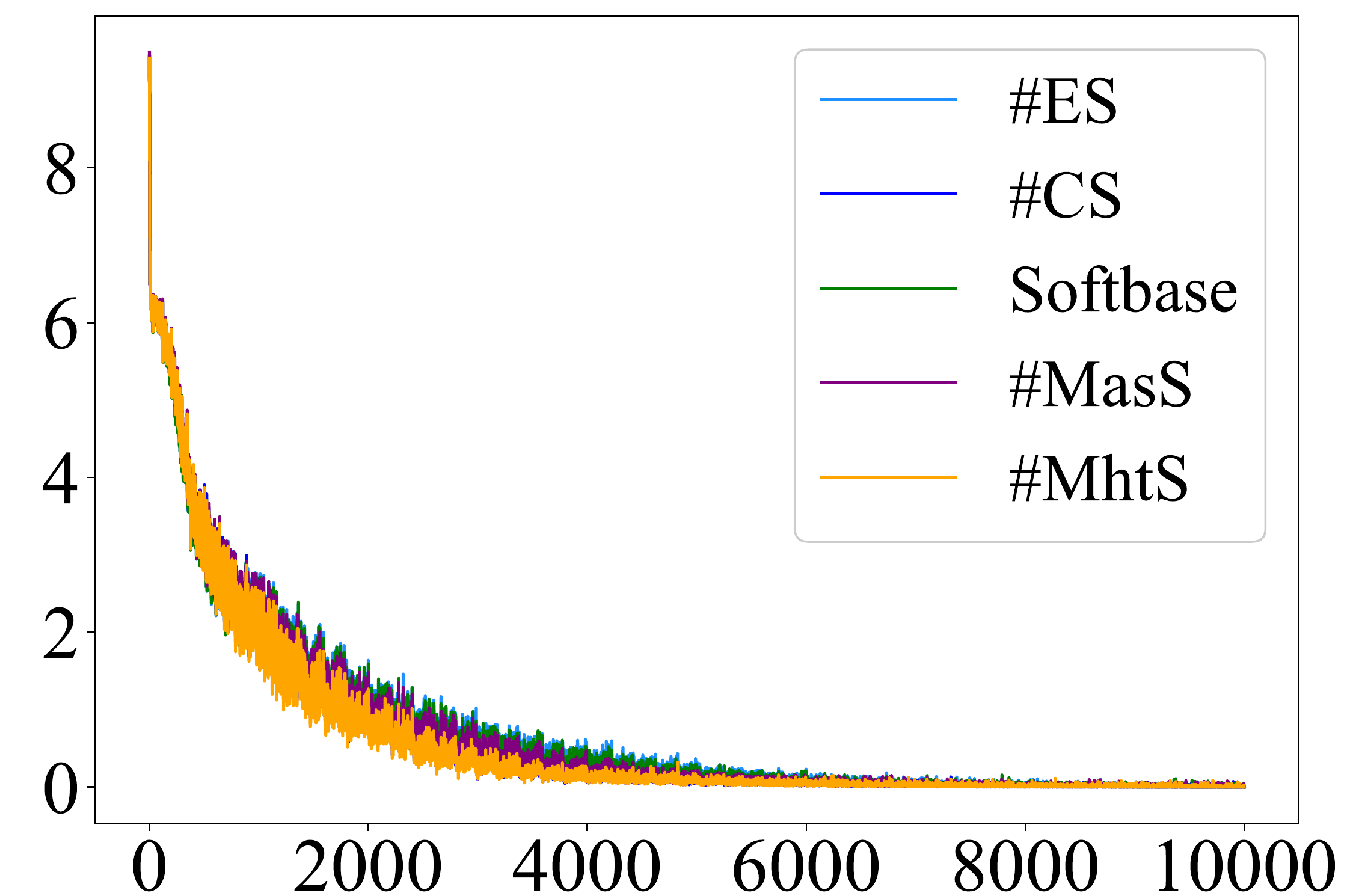}}
      \label{fig:side:b}   
  \end{minipage}
 \begin{minipage}[t]{0.245\linewidth} 
  \centering
  \subfigure[Hard SSM on WHG]{
    \includegraphics[width=1.4in]{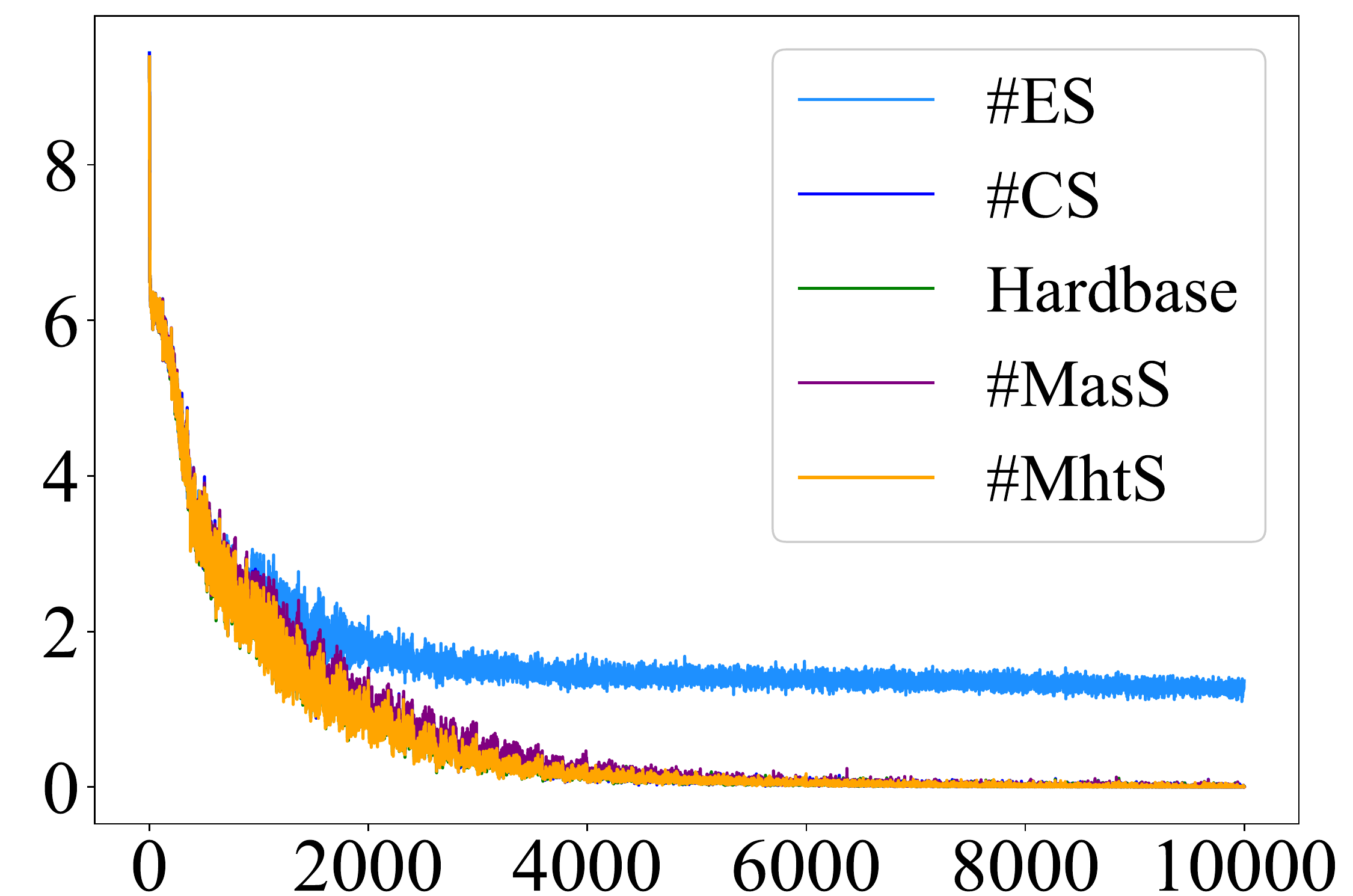}}
      \label{fig:side:a}   
  \end{minipage}%
  \begin{minipage}[t]{0.245\linewidth}   
      \centering   
      \subfigure[Soft SSM on THG]{
  \includegraphics[width=1.4in]{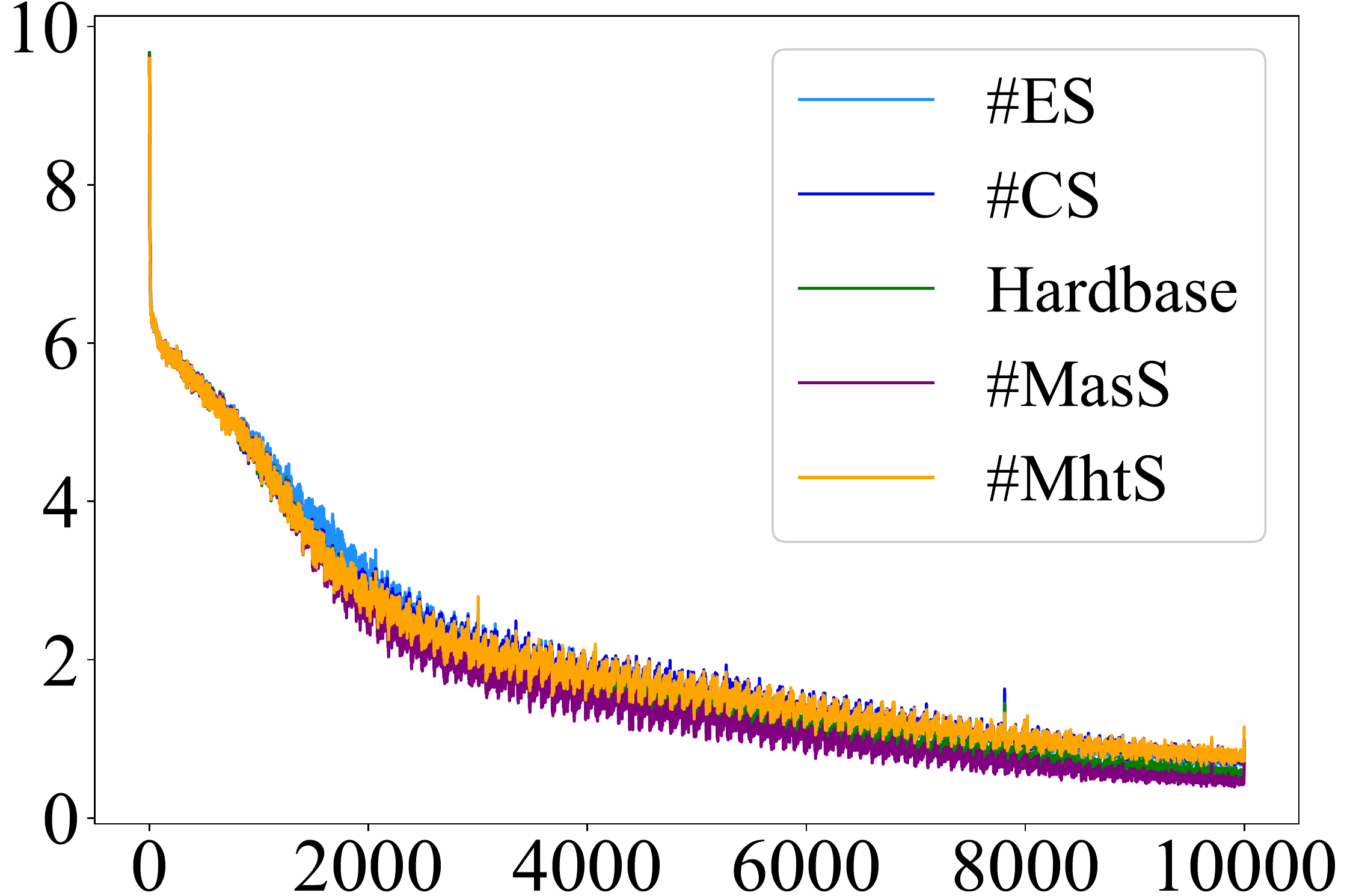}}
      \label{fig:side:b}   
  \end{minipage}
  \begin{minipage}[t]{0.245\linewidth}   
   \centering   
   \subfigure[Hard SSM on THG]{
 \includegraphics[width=1.4in]{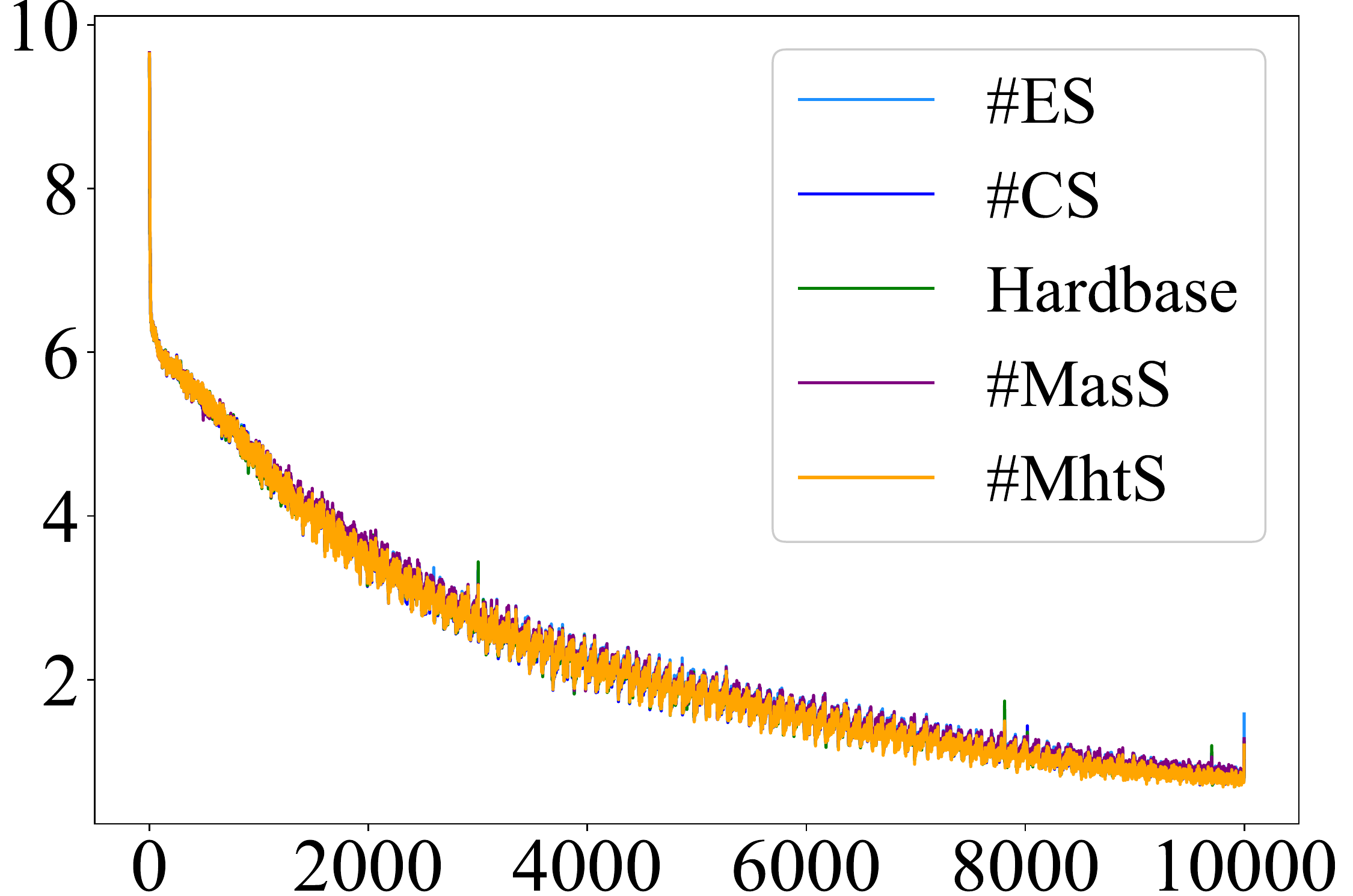}}
   \label{fig:side:b}   
 \end{minipage}
    \caption{The training loss and performance in different SSM on two datasets.}
    \label{curves}
 \end{figure*}

\begin{figure*} 
  \centering
  \begin{minipage}[t]{0.245\linewidth} 
  \centering
  \subfigure[Soft SSM +C on WHG]{
    \includegraphics[width=1.45in]{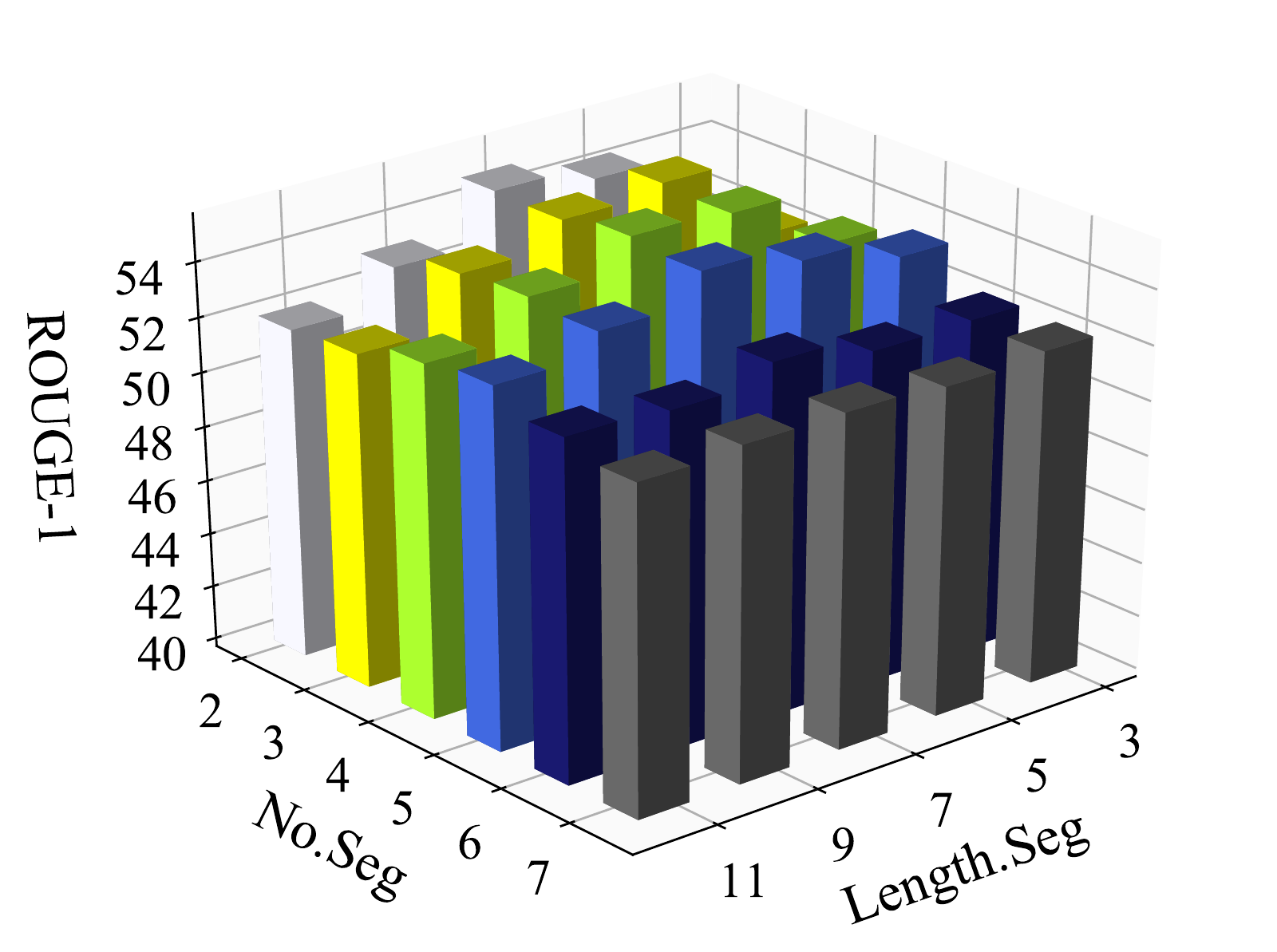}}
  \end{minipage}%
  \begin{minipage}[t]{0.245\linewidth}   
  \centering   
      \subfigure[Hard SSM +C on WHG]{
  \includegraphics[width=1.45in]{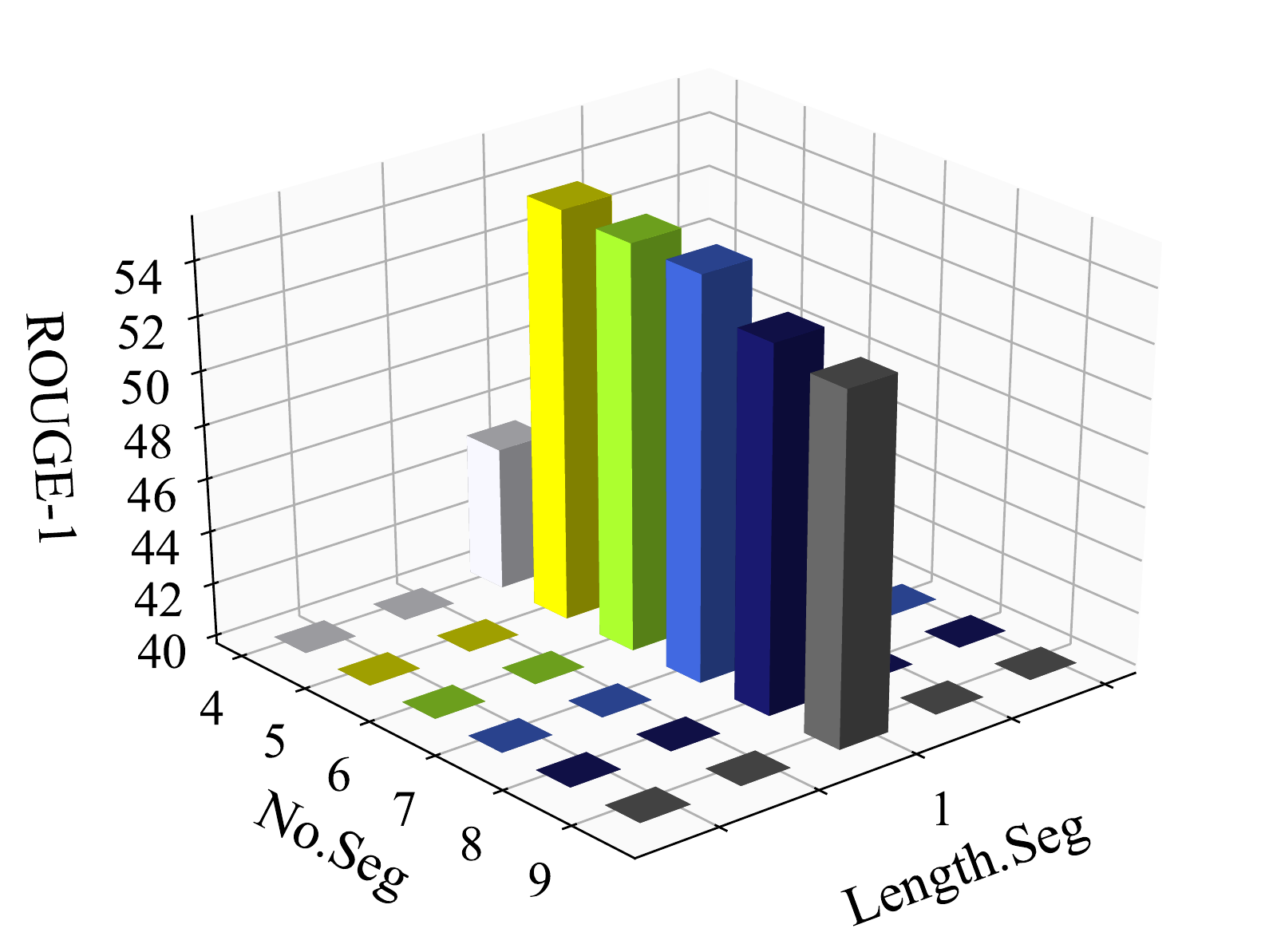}}
  \end{minipage}
    \begin{minipage}[t]{0.245\linewidth}   
  \centering   
      \subfigure[Soft SSM on THG]{
  \includegraphics[width=1.45in]{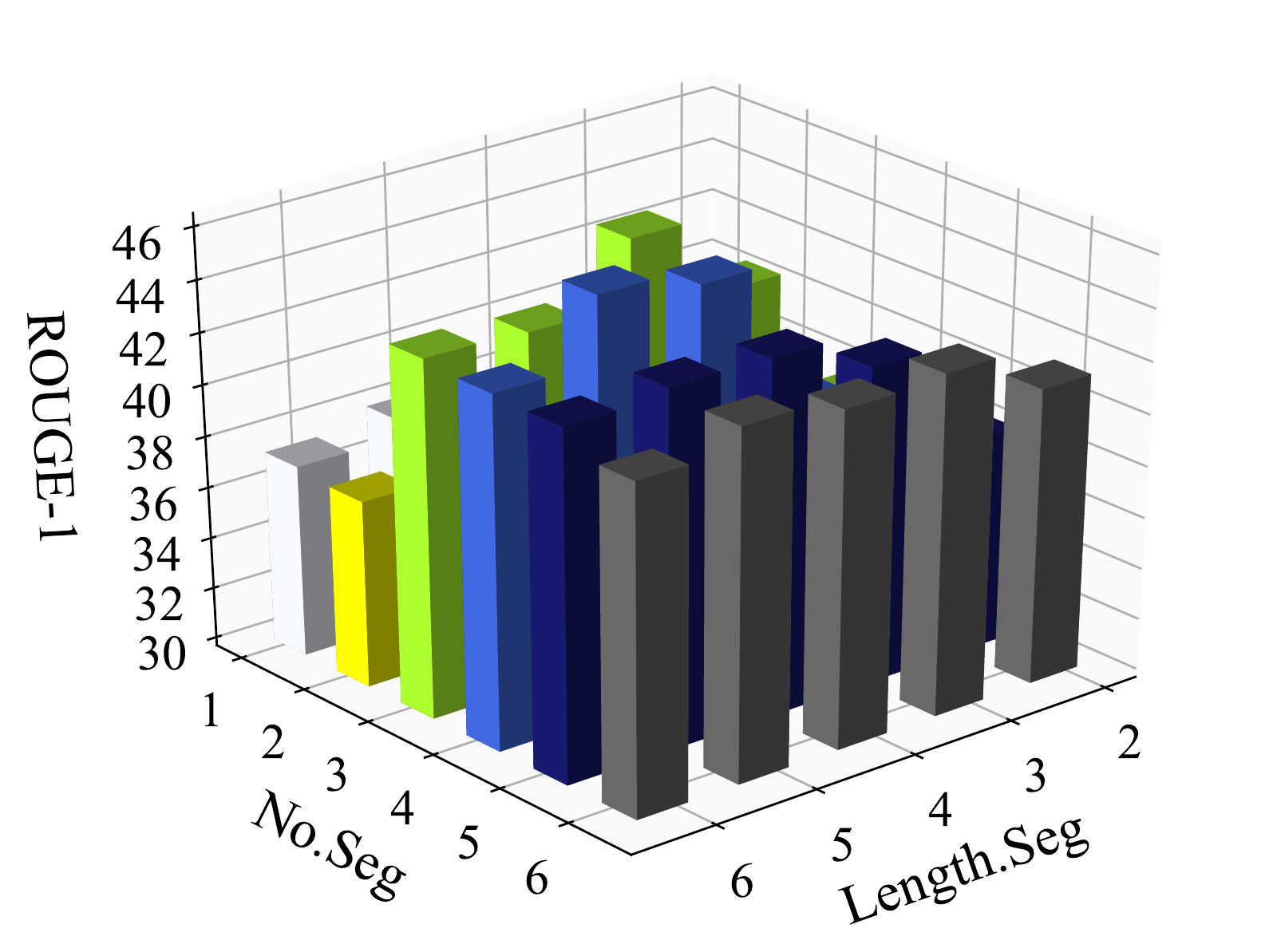}}
  \end{minipage}
    \begin{minipage}[t]{0.245\linewidth}   
  \centering   
      \subfigure[{Hard SSM on THG}]{
  \includegraphics[width=1.45in]{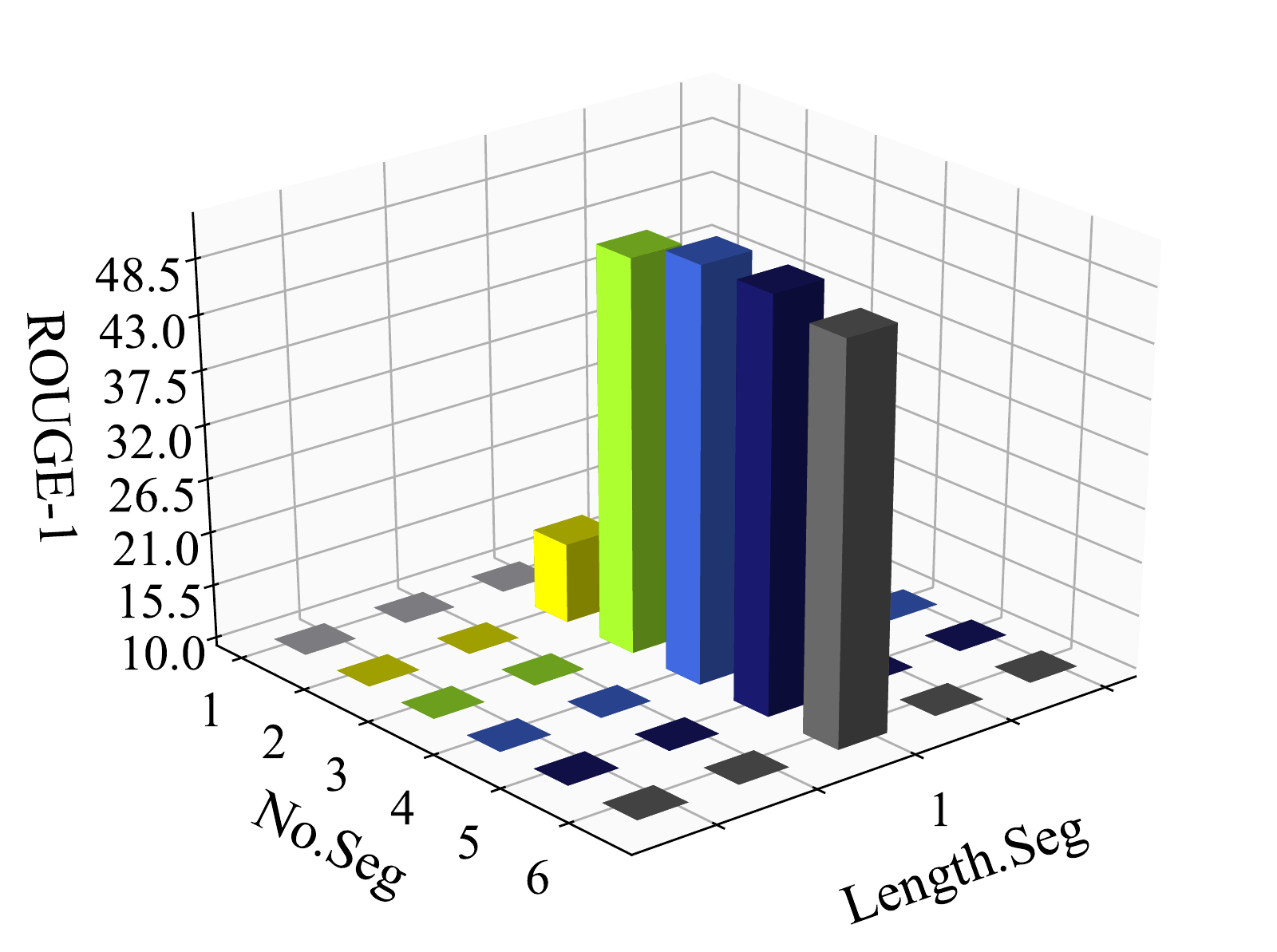}}
  \end{minipage}
    \caption{Hyperparameters searching for the number of selected segments (No.Seg) and length of segments (Length.Seg) during evaluation of ROUGE-1 performance. The Hard SSM only selects the [SEG] token, so the length of segments is 1.  
    }
    \label{segmentationresults}
\end{figure*}

  

  

\section{Results and Analysis}
\subsection{Main Results} 
As shown in Table~\ref{mainresults}, we release results on Weibo and Twitter hashtag generation datasets separately. 
Our S{\footnotesize EG}T{\footnotesize RM} (soft) consistently gets the best performance on both datasets. 
Its hard-based version is superior to the Softbase model. 
However, hard-based selection models are invalidated on the Twitter dataset. 
The reason may be that the text of the Twitter posts is too short.
The average length of the original text of the Twitter posts is 23, which is far less than 75 of the Weibo posts.
In the Twitter posts, each segment attends minor content, which is not conducive to hashtag generation.

According to F1@\(k\) scores, we find it is difficult for Twitter to normalize the hashtags.
The reason probably is that some hashtags are low-frequency words or abbreviations.
Those rare hashtags (less frequent in distribution) are not fully trained, resulting in low F1 scores for English Twitter. 
Moreover, the models' efficiency about the R{\footnotesize OUGE}-2 on the THG dataset is worse than that of the WHG dataset. 
The reason for it may be that English Twitter contains massive abbreviated and single-word hashtags, which causes the insufficient training of those low-frequency hashtags.

\noindent\textbf{Comparison with keywords generation method}. 
Compared with keywords extraction method TFIDF and generation method ExHiRD, our model S{\footnotesize EG}T{\footnotesize RM} (soft) obtains significant improvements on most of the metrics (paired t-test, p \(< \) 0.05). 
Besides, ExHiRD has inherent defects, such as insufficiency of long-term sequence dependency.
Another serious drawback is that these models are hard to generate phrase-level Weibo hashtags.

We conclude that keywords extraction methods hardly adapt to large-scale datasets since they can not reorganize words appropriately. 
The reason may be that most hashtags only appear a few times. 
Given such a large and imbalanced hashtag space, hashtag selection from a candidate list,  might not perform well. 
This conclusion is consistent with the findings of ~\citet{WangLKLS19} and ~\citet{Zhang19}.

\noindent\textbf{Comparison with selective encoding systems}. 
Whether testing in R{\footnotesize OUGE} or F1@K, we find that our selective model always appears to be better than S{\footnotesize EASS} and B{\footnotesize OTTOM}U{\footnotesize P} with a certain margin. 
S{\footnotesize EASS} fails in its long-term semantic dependence. 
B{\footnotesize OTTOM}U{\footnotesize P} fails in its complex joint optimization on two objectives of word selection and generation.

\noindent\textbf{Comparison with salient Transformer generator}. 

Among those Transformer-based models, our S{\footnotesize EG}T{\footnotesize RM} is superior to the selective encoding model B{\footnotesize OTTOM}U{\footnotesize P} and the salient T{\footnotesize RANS}A{\footnotesize BS}. 
The superiority of our model can be attributed to the explicit selection of dominant pieces and modeling of the segmental compositionality.


\subsection{Comparison of Different SSM}
\label{CDS}
\noindent\textbf{Performance of SSMs}. 
Firstly, as shown in Table~\ref{mainresults}, the results of ablation experiments (without any SSM) are to compare base models (S{\footnotesize EG}T{\footnotesize RM} Softbase) with soft S{\footnotesize EG}T{\footnotesize RM}.
The results indicate the superiority of SSM.
Secondly, the results of different SSMs can be seen in Table~\ref{selectionresults}. 
To simplify the description, we use `\#SSM' to simply represent a method. 
\#ES always gets the worst performance, whether for hard or soft segments selection, which indicates a poor segment selection will produce adverse input  (e.g., it does not pick out critical information) to the decoder, making a worse generation.  
For Twitter hashtag generation, \#C, \#MasS, and \#MhtS outperform the corresponding models of Softbase or Hardbase.
\#MhtS is the best among hard-based SSM models, and \#MasS is the best among soft-based SSM models. 
Observing the loss curves in Figure~\ref{curves}.(e) and Figure~\ref{curves}.(f) and R{\footnotesize OUGE} F1 results on Table~\ref{selectionresults}, we find that the lower the convergence loss is, the more R{\footnotesize OUGE} can be obtained. 

\noindent\textbf{Hyperparameter searching for SSMs}. 
To search for an optimal number of segments, we compare different SSMs on the hyperparameter of Top \(k\) segments selection in Figure~\ref{segmentationresults}.
Compared with the baselines, the performance of different selected segments for hard soft can be guaranteed among [2,7] on WHG and [3,6] on THG. 
The number can be [5,9] on WHG, [3,6] on THG for soft SSM. 

We can also analyze the length of the soft SSM to discourse how many selected tokens are appropriate in segments from the  Figure~\ref{segmentationresults}.
For WHG, the superior length of segments is [5,9], and for THG, it is [3,6]. 
Besides, it can be roughly observed that the model often obtains good performance when both the number of segments and the length of segments are not too large or too small, especially for the soft SSM. Too few segments make the model lack valuable information, resulting in a steep decline in performance. Too long segments will bring redundant information, making it equal to the performance of the base Transformer model.

These results also indicate that hashtags are mostly assembled from scattered semantic pieces, and attending to those key segments can distill unnecessary information and stabilize performance.

\subsection{N-gram Overlaps}
\begin{figure*}[width=2.1\linewidth,cols=4,pos=h]  
   \begin{minipage}[t]{0.245\linewidth}   
     \centering   
      \subfigure[Soft SSM on WHG]{
 \includegraphics[width=1.4in]{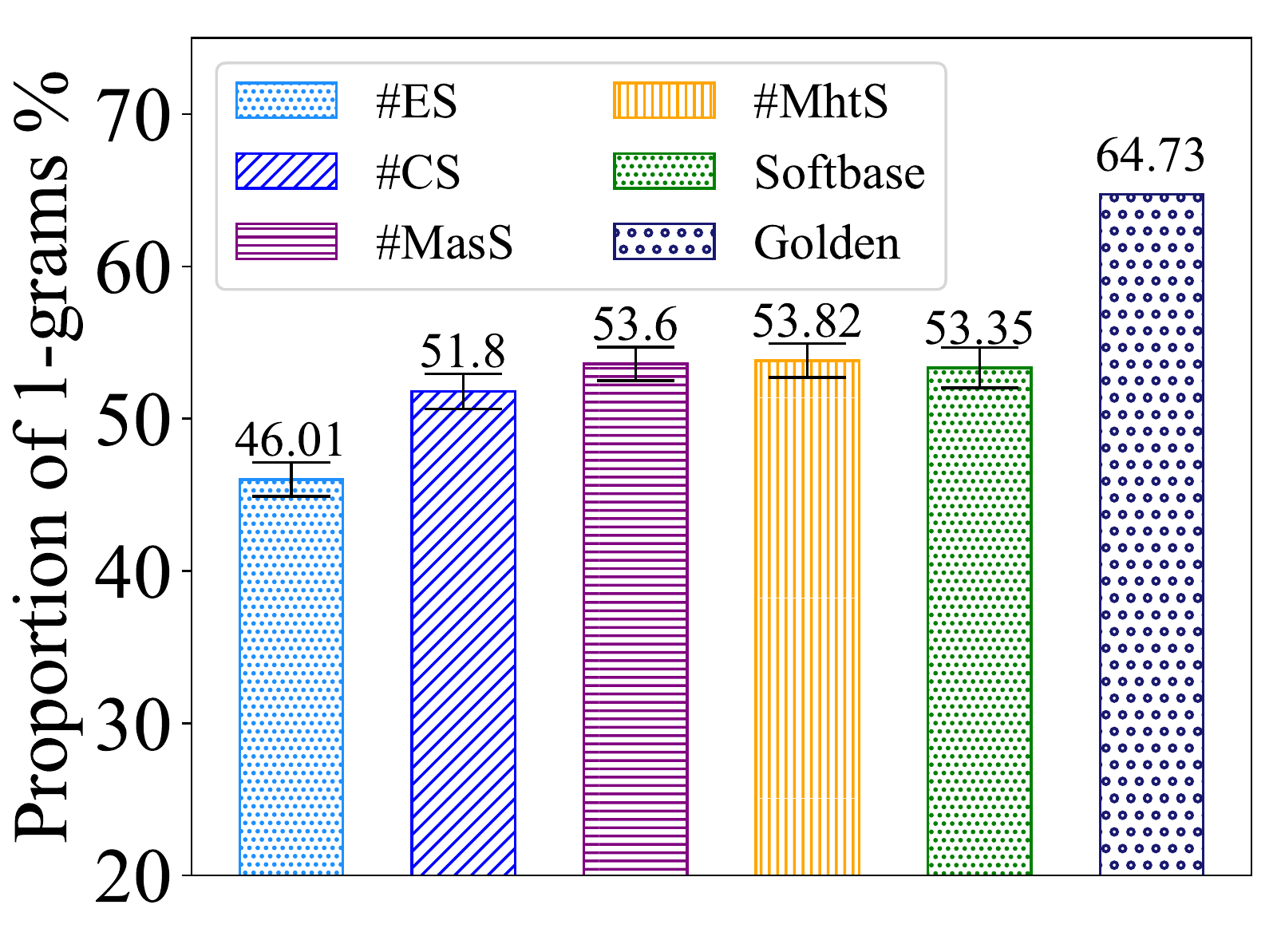}}
     \label{fig:side:c}   
 \end{minipage}
 \begin{minipage}[t]{0.245\linewidth} 
   \centering
       \subfigure[Soft SSM on WHG]{
   \includegraphics[width=1.4in]{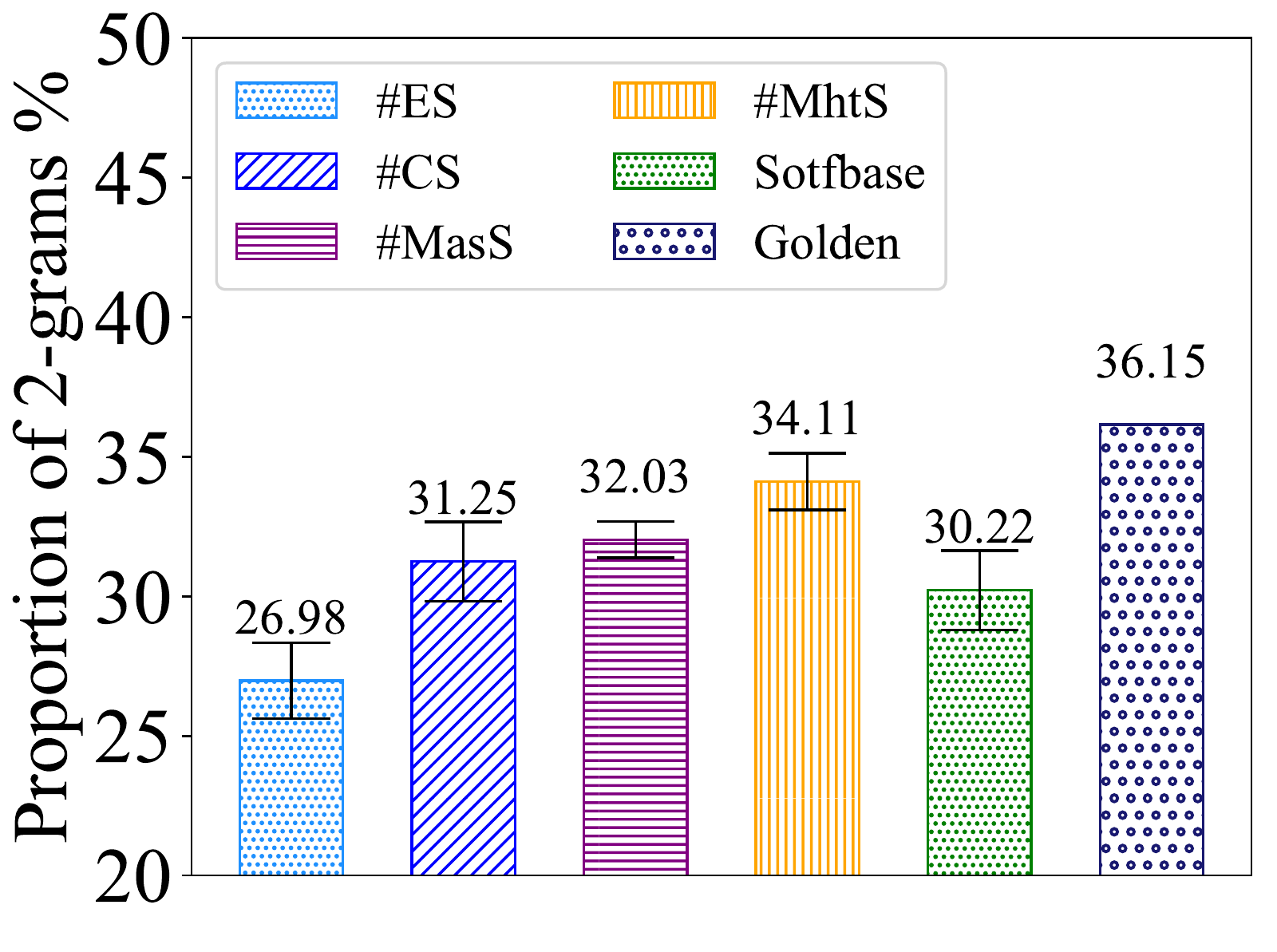}}
       \label{fig:side:d}   
   \end{minipage}%
 \begin{minipage}[t]{0.245\linewidth} 
   \centering
       \subfigure[Hard SSM on WHG]{
   \includegraphics[width=1.4in]{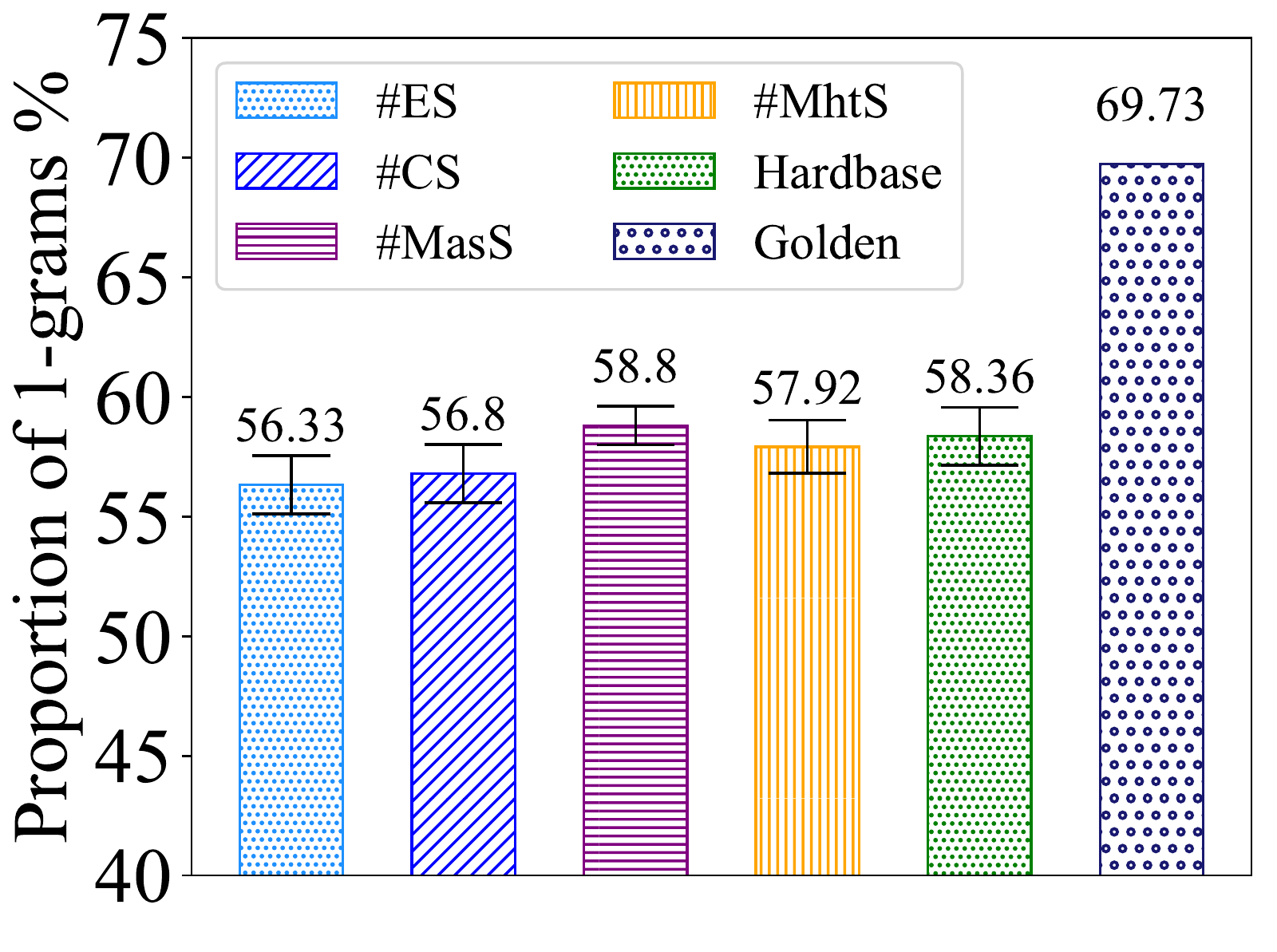}}
       \label{fig:side:g}   
   \end{minipage}%
   \begin{minipage}[t]{0.245\linewidth}   
       \centering   
        \subfigure[Hard SSM on WHG]{
   \includegraphics[width=1.4in]{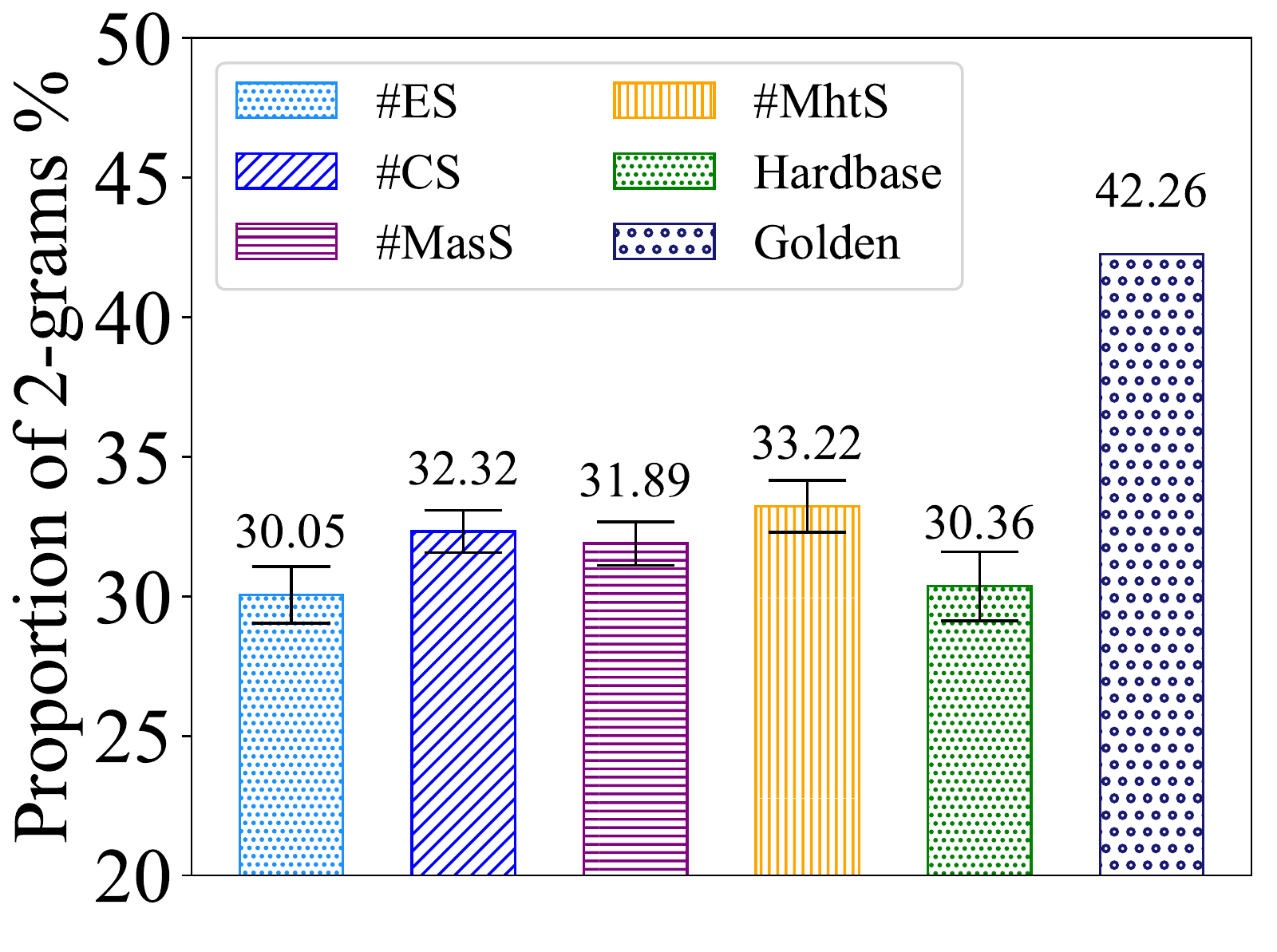}}
       \label{fig:side:h}   
   \end{minipage}

   \begin{minipage}[t]{0.245\linewidth}   
     \centering   
      \subfigure[Soft SSM on THG]{
 \includegraphics[width=1.4in]{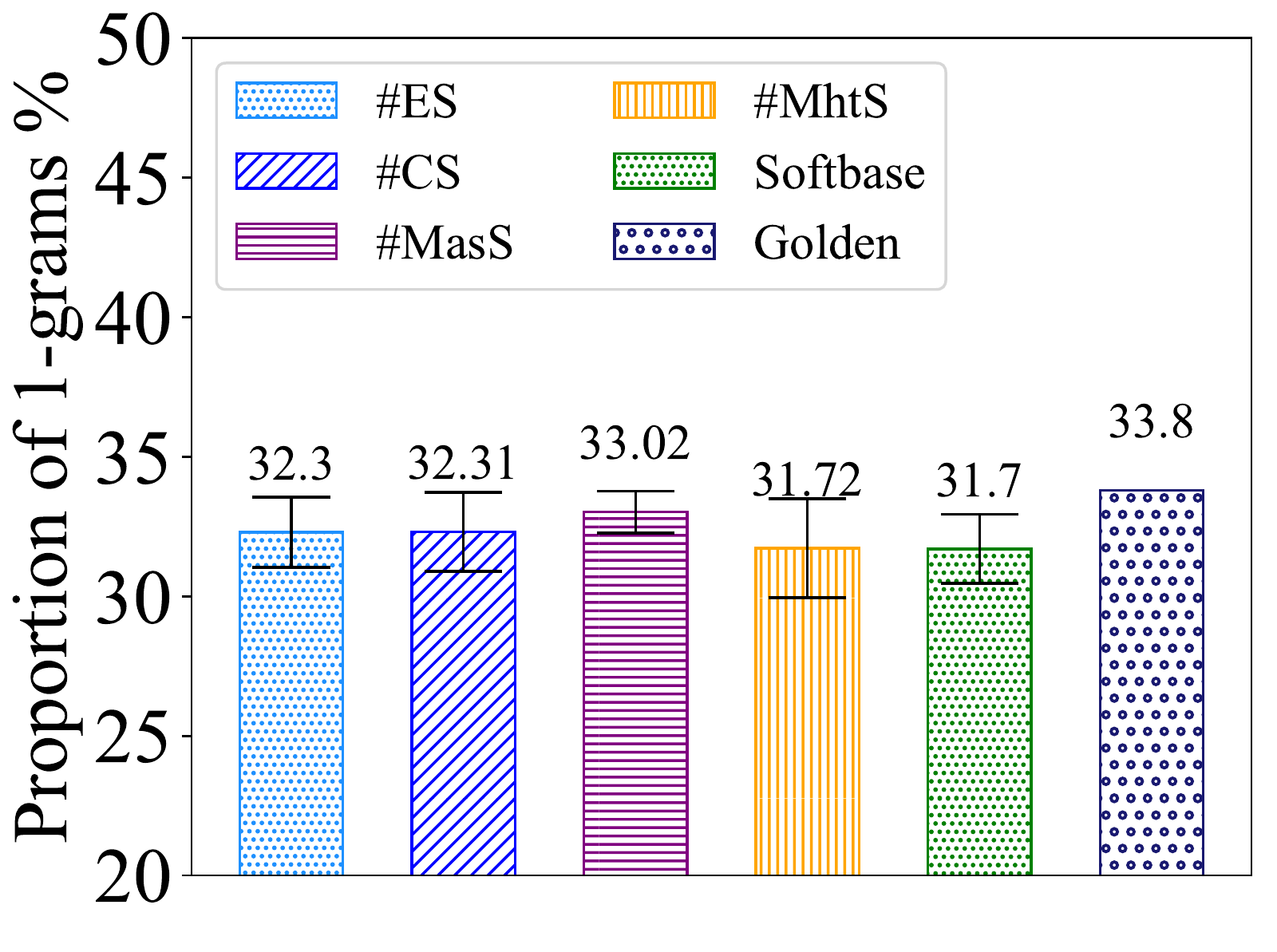}}
     \label{fig:side:i}   
 \end{minipage}
 \begin{minipage}[t]{0.245\linewidth} 
   \centering
       \subfigure[Soft SSM on THG]{
   \includegraphics[width=1.4in]{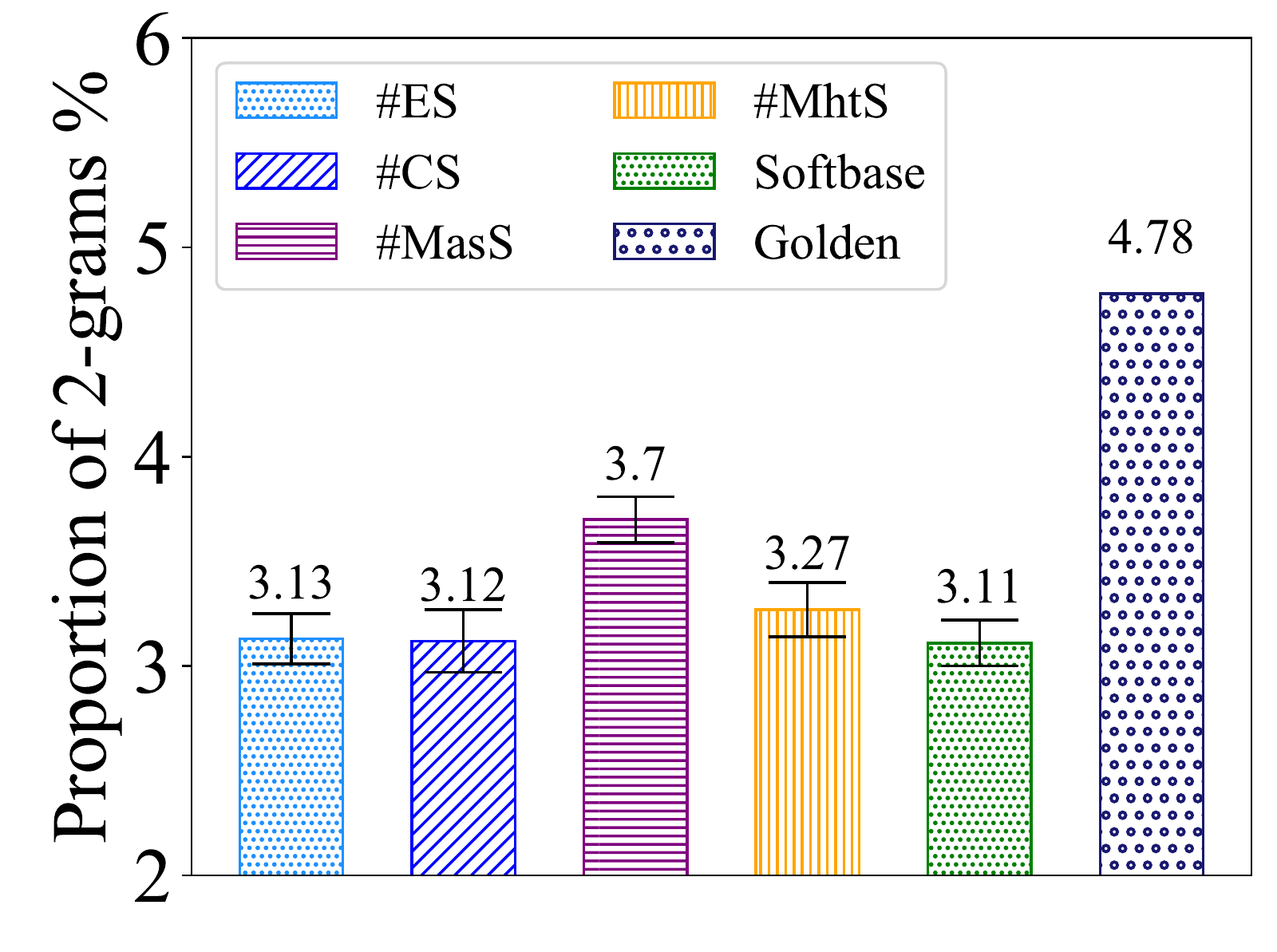}}
       \label{fig:side:j}   
   \end{minipage}%
   \begin{minipage}[t]{0.245\linewidth}   
       \centering   
        \subfigure[Hard SSM on THG]{
   \includegraphics[width=1.4in]{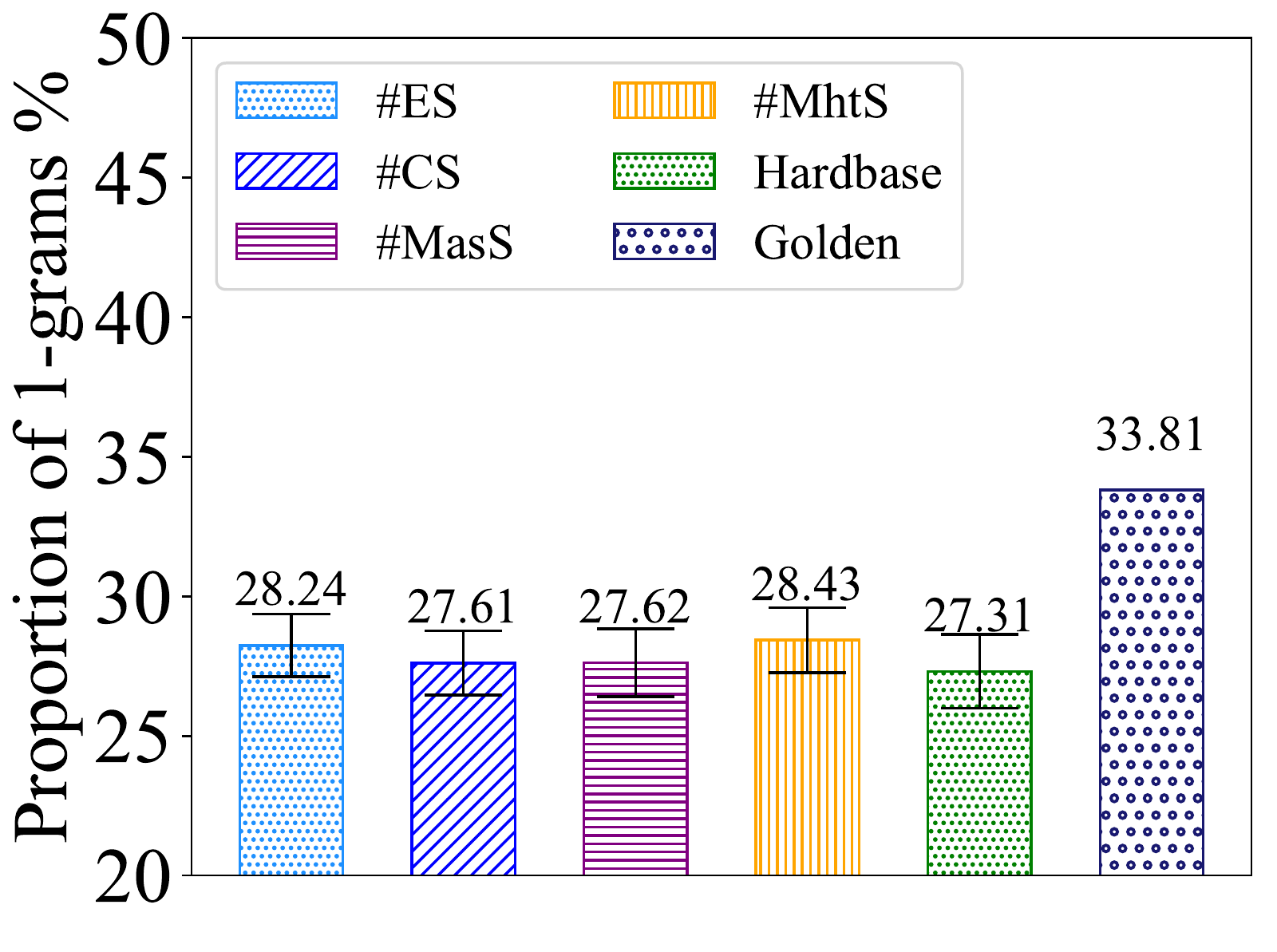}}
       \label{fig:side:k}   
   \end{minipage}
   \begin{minipage}[t]{0.245\linewidth}   
     \centering   
      \subfigure[Hard SSM on THG]{
 \includegraphics[width=1.4in]{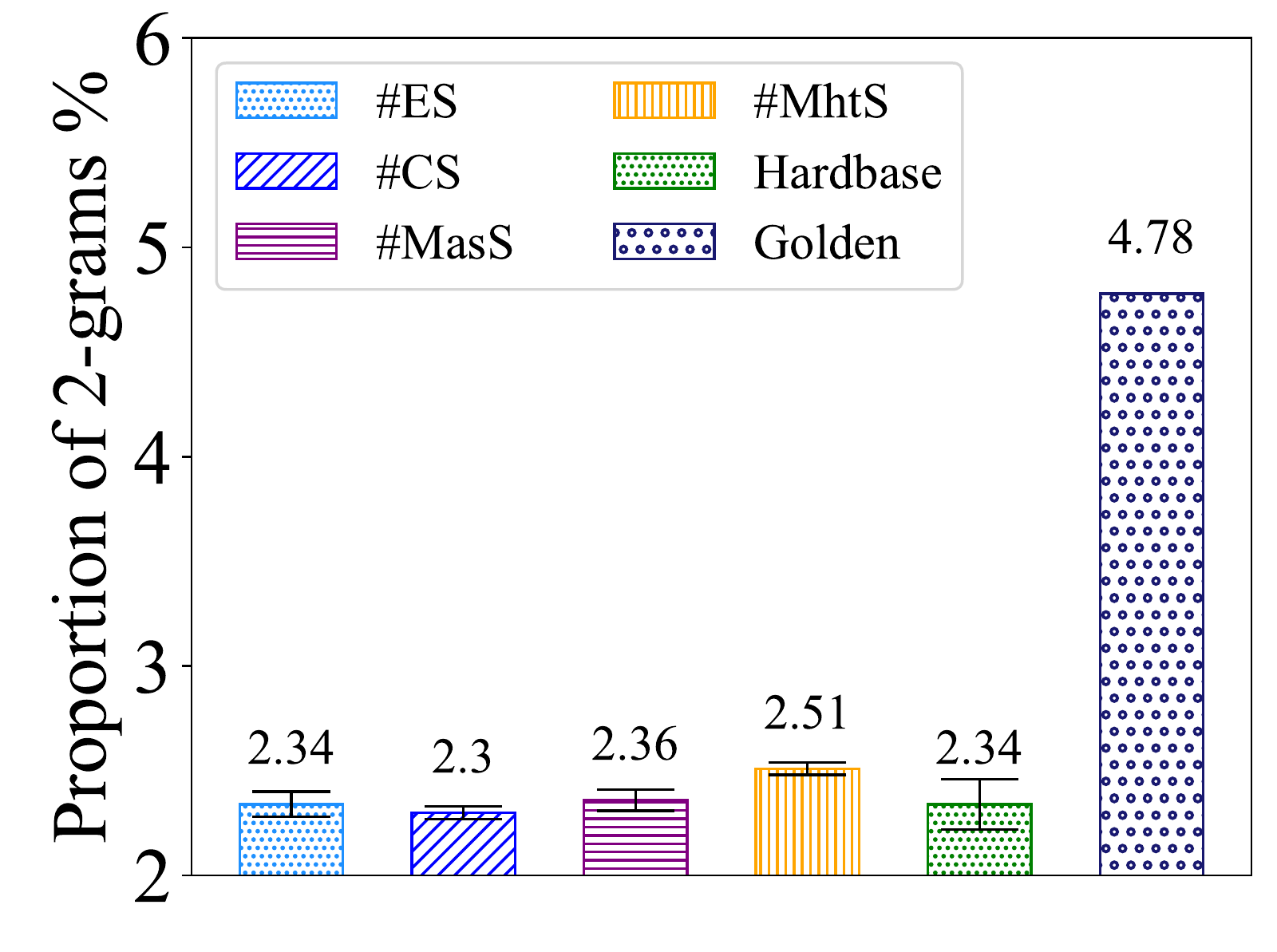}}
     \label{fig:side:l}   
 \end{minipage}
     \caption{N-gram overlaps of the generated summaries and the golden summaries to their corresponding source articles. 
     The results are $means \pm S.D.\left ( n=3\right )$.
       }
     \label{ngram}
 \end{figure*}

To test the extraction ability of our systems, we illustrate the comparison of n-gram overlaps for our models.
In Figure~\ref{ngram}, the generated hashtags of \#MhtS overlap the post text more often than other selection methods on the WHG dataset.  
As shown in Figure~\ref{ngram} (a). 59.42\% of the generated 1-gram is duplicated from the posts' 1-gram). 
For the proportion of 2-gram overlaps, as shown in Figure~\ref{ngram} (b), \#MhtS is almost close to golden hashtags, with only tiny differences.

The results indicate that our model can duplicate tokens from the source text and simultaneously retain accuracy. 
Our segment selection mechanism makes the system more reliable to reorganize key details correctly. 
Almost all soft SSMs, except for \#ES, rewrite substantially more abstractive hashtags than our base model, which has no segment selection mechanism. 
Our segment selection model allows the network to copy words from the source text and consults the language model simultaneously to extract words from vocabulary, enabling operations like truncation and stitching to be performed accurately.

 \section{Conclusion and Future work}
To generate microblog hashtags automatically and effectively in a large-scale dataset, we modify the Transformer's deep architecture by integrating a segments-selection module. The model is aware of filtering out secondary information under different granularity among text, segments, and tokens.  
The experimental results validated in two constructed large-scale datasets indicate that our model achieves state-of-the-art effects with significant improvements. We will apply an indefinite length segmentation scheme in future works, such as indefinite clause-based segment-selection for generation. The method of pre-training language model is also worthy for exploration.

\begin{CJK}{UTF8}{gkai}
  \begin{table*}[width=2.1\linewidth,cols=4,pos=h] 
    \scriptsize
    \caption{Four cases of the generated hashtags for Weibo posts and a generation case for Twitter posts. 
    The last two cases in Chinese are carefully translated as shown in brackets for the convenience of reading and comparison. 
    }
      \renewcommand\arraystretch{1.2}
    \begin{tabular}{@{}|p{17cm}|@{}}
    \hline
    \textbf{The Twitter post for hashtag generation}： 
    We're re-opening our Helen Albert Certified Farmers' Market on Monday, September 14 from 9 AM to 2 PM with new safety measures in effect. The Farmers' Market features organic and farm fresh fruits and vegetables, baked goods, fresh fish, and more.        
    \\ 
    \textbf{Golden}: \# organic farmers market     \\ 
    \textbf{S{\scriptsize EG}T{\scriptsize RM} {\scriptsize Softbase}}: \# farmers market     \\ 
    \textbf{S{\scriptsize EG}T{\scriptsize RM} {\scriptsize Hardbase}}: \# farmers \#market   \\ 
    \textbf{S{\scriptsize EG}T{\scriptsize RM}} (hard): \# organic farmers \#market  \\ 
    \textbf{S{\scriptsize EG}T{\scriptsize RM}} (soft): \# organic farmers market  \\ 
    \hline
    \hline
    \textbf{The Twitter post for hashtag generation}： 
    An event organized by the Italian Presidency of G20, UNDP and  UNEP, with the contribution of Urban 20 focused on multi-level governance aspects of Nature-based solutions in cities. 
    \\ 
    \textbf{Golden}: \# G20 Italy \# U20          \\ 
    \textbf{S{\scriptsize EG}T{\scriptsize RM} {\scriptsize Softbase}}: \# G20 Italy  \# G20           \\ 
    \textbf{S{\scriptsize EG}T{\scriptsize RM} {\scriptsize Hardbase}}: \# G20      \\ 
    \textbf{S{\scriptsize EG}T{\scriptsize RM}} (hard): \# G20 Italy \# Urban 20  \\ 
    \textbf{S{\scriptsize EG}T{\scriptsize RM}} (soft): \# G20 Italy \# Urban 20  \\ 
    \hline
    \hline
    \textbf{The Weibo post for hashtag generation}： 
    9月3日下午在厦门召开的金砖国家工商论坛开幕式上，国家主席习近平发表题为《共同开创金砖合作第二个“金色十年”》的主旨演讲。
    
    On the afternoon of September 3rd, at the opening ceremony of the BRIC industrial and commercial forum held in Xiamen, President Xi Jinping delivered a Keynote speech entitled Jointly Creating the Second  `Golden Decade' of BRICS Cooperation.         
    \\ 

    \textbf{Golden} : \# \underline{金}  \underline{砖}  \underline{厦}   \underline{门}  \underline{会}\underline{晤} \#  (\#Meetings of BRICS in Xiamen\#) \\ 
    \textbf{S{\scriptsize EG}T{\scriptsize RM} {\scriptsize Hardbase}} : \# \underline{金}  \underline{砖}  \underline{耀}  \underline{鹭}  \underline{岛} \# (\#Meetings of BRICS at Banlu Island\#) \\
    \textbf{S{\scriptsize EG}T{\scriptsize RM} {\scriptsize Softbase}} : \# \underline{金}  \underline{砖}  \underline{会}  \underline{议}  \# (\#Meetings of BRICS\#) \\
    \textbf{S{\scriptsize EG}T{\scriptsize RM}} (\& hard): \# \underline{金}  \underline{砖}  \underline{厦}  \underline{门}  \underline{会} \underline{晤} \# (\#Meetings of BRICS in Xiamen\#) \\
    \textbf{S{\scriptsize EG}T{\scriptsize RM}} (\& soft): \# \underline{金}  \underline{砖}  \underline{厦}  \underline{门}  \underline{会} \underline{晤} \# (\#Meetings of BRICS in Xiamen\#) \\
    \hline
    \hline
    \textbf{The Weibo post for hashtag generation}： 
    全场比赛结束，巴塞罗那主场5:0战胜西班牙人赢得本赛季
    首场同城德比，梅西上演帽子戏法，皮克、苏亚雷斯锦上添花，拉基蒂奇和阿尔巴分别贡献两次助攻，登贝莱首秀助攻苏亚雷斯。
    
    At the end of the match, Barcelona defeated the Spaniards 5-0 at home to win the first derby in the same city this season. Messi staged a hat-trick, Pique and Suarez were icing on the cake, Rakitic and Alba contributed two assists, Dembele assisted Suarez in his first match.       
    \\ 
    \textbf{Golden} : \# \underline{巴}  \underline{塞}  \underline{罗}   \underline{那}  \underline{vs.} \underline{西} \underline{班} \underline{牙} \underline{人}\# (\#Barcelona vs. Spaniards\#)   \\ 
    \textbf{S{\footnotesize EG}T{\footnotesize RM} {\scriptsize Hardbase}} : \# \underline{巴}  \underline{塞}  \underline{罗}   \underline{那}  \underline{vs.} \underline{尤} \underline{文} \underline{图} \underline{斯} \# (\#Barcelona vs. Juventus\#)  \\
    \textbf{S{\footnotesize EG}T{\footnotesize RM} {\scriptsize Softbase}} : \# \underline{巴}  \underline{塞}  \underline{罗}   \underline{那}  \underline{vs.} \underline{西} \underline{班} \underline{牙} \underline{人} \# (\#Barcelona vs. Spaniards\#)  \\
    \textbf{S{\footnotesize EG}T{\footnotesize RM}} (hard):  \# \underline{巴}  \underline{塞}  \underline{罗}   \underline{那}  \underline{vs.} \underline{西} \underline{班} \underline{牙} \underline{人} \# (\#Barcelona vs. Spaniards\#) \\ 
    \textbf{S{\footnotesize EG}T{\footnotesize RM}} (soft):  \# \underline{巴}  \underline{塞}  \underline{罗}   \underline{那}  \underline{vs.} \underline{西} \underline{班} \underline{牙} \underline{人} \# (\#Barcelona vs. Spaniards\#) \\ 
    \hline
    \end{tabular}
    \label{fig:Case}
    \vspace{-0.2in}
\end{table*}
\end{CJK}

\appendix
\section{Appendix}
\subsection{Dataset Construction}
\label{DC}

Most of the posts are attached with informative hashtags in the Social Networking Services (SNS) platform. 
In Figure~\ref{hashtag}, the microblog user has posted a hashtag `\textit{\#5G Bring New Value}', and we treat such natural user-provided hashtags as ground-truth for training, validation, and test.  
Besides, post-hashtags are selected by users such as official media and influencers whose labeled hashtags are of high quality. 
These premises make it reasonable to directly use the user-annotated hashtags in microblog as the ground-truth hashtags.
We take the user-annotated hashtags appearing in the beginning or end of a post as the reference as the prior  works~\cite{ZhangWGH16,ZhangLSZ18,WangLKLS19} did~\footnote{Hashtags in the middle of a post are not considered as they generally act as semantic elements rather than topic words.}.

\begin{table}[tb]
 \centering
 \footnotesize
 \caption{Statistics of the hashtags. $\mathcal{N_{W}}$: the percentage of hashtags existing new words that do not appear in a post text. $\mathcal{P}$: the percentage of hashtags composed of 1 or 2 words. $\mathcal{S}$: the percentage of hashtags in which the words appear in different segments of a post text. 
$\mathcal{N}$: the maximum number of the segments that contain words from hashtags.}
 \renewcommand\arraystretch{1}
 \setlength{\tabcolsep}{4.6mm}{
   \begin{tabular}{@{}l|cccc}
 \toprule
   Datasets          &  $\mathcal{N_{W}}$  & $\mathcal{P}$  & $\mathcal{S}$  & $\mathcal{N}$  \\ 
 \midrule
WHG                  
& 10.32\% & 1.54\% & 61.63\% & 15  \\
THG            
& 10.20\% & 60.43\% & 15.31\% & 4  \\
\bottomrule
\end{tabular}}
\label{Statistics}
\end{table}

\textbf{WHG construction}: We collect the post-hashtag pairs by crawling the microblog of seed accounts involving multiple areas from Weibo. 
These seed accounts, such as \textit{People's Daily}, \textit{People.cn}, \textit{Economic Observe press}, \textit{Xinlang Sports}, and other accounts with more than 5 million followers come from different domains of politics, economic, military, sports, etc. 
The post text and hashtag pairs are filtered, cleaned, and extracted with artificial rules.  
We remove those pairs with too short text lengths (less than 60 characters) that only account for a small part of all data. 
Statistics of the WHG datasets are shown in Table ~\ref{Statistics}, in which about 10.32\% of the hashtags contain new words that do not appear in a post text, and 61.63\% of the hashtags have words that appear in three or more different segments. 
At most, 15 segments contain words from hashtags.

\textbf{THG construction}: We use TweetDeck~\footnote{\url{https://tweetdeck.twitter.com/}} to get and filter tweets. 
We collect 200 seed accounts, such as organizations, media, and other official users, to obtain high-quality tweets. 
Then the Twitter post-hashtag pairs are crawled from the seed users.
The tokenization process is integrated into the training, evaluating, and testing step. 
We use RoBERTa's FullTokenizer and vocabulary~\cite{devlin2018bert}. 

Table~\ref{Statistics} shows that about 10.20\% of the hashtags contain new words that do not appear in posts. 
About 28.41\% of the hashtags consist of a single word or abbreviation.  
At most, 4 segments contain words from hashtags.
We employ 204,039 post-hashtag pairs for training in the THG dataset, 11,335 and 11,336 pairs for validation and test, respectively.

\subsection{Case Study}
  \label{sec:Casestudy} 
  We illustrate some generated hashtags of our implemented models for Chinese Weibo and English Twitter. 
  As indicated by the examples in Table~\ref{fig:Case}, all generated hashtags have pinpointed the core meaning of posts with fluency. 
  Hashtags are truncated to form shorter versions and are composed of discontinuous tokens. 
 Comparing the generations of our models to Golden hashtag, we find two base models and hard-based S{\footnotesize EG}T{\footnotesize RM} have generated usable hashtags (e.g., `\#farmers', `\#market', and `\#organic farmers') which are duplicated tokens from the source text.
Although these hashtags are not the same as the Golden ones (resulting in a low F1 value), they are usable.
  This case also proves the reason why we choose R{\footnotesize OUGE} evaluation, namely n-gram overlaps will not miss the highly available hashtags.

 Hashtags generated by S{\footnotesize EG}T{\footnotesize RM} are almost entirely consistent with the Golden ones.
 For English twitter, it is not easy to generate hashtags or abbreviations.
 For example, in case 2, there are two hashtags, where `\textit{U20}' refers to `\textit{Urban 20}' in the original text. 
 Our S{\footnotesize EG}T{\footnotesize RM} directly selects the phrase `\textit{Urban 20}' in the original text as the generation result. 
 This is an apparent correct hashtag but results in a comparative low R{\footnotesize OUGE} score. 
 This case also shows the necessity of using n-gram overlaps to evaluate the performance, which will not omit the case of choosing the correct phrase as a hashtag from the original text. 

  In the third case of Weibo hashtag generation, the Hardbase generates `\textit{BRICS in Yaolu island}'. 
 Although two generated results are not identical to the Golden ones, they are all correct facts.
 For example, `\textit{Yaolu}', an island of `\textit{Xiamen}' is the specific location of the `\textit{BRICS conference}'.
  There exists a wrong generation in the fourth case.
  Hashtags generated by Hardbase of S{\footnotesize EG}T{\footnotesize RM} contain an unrelated term of `\textit{Juventus}' which is a club in Italy. 

\subsection{Limitations}
Our S{\scriptsize EG}T{\scriptsize RM} significantly boosts the performance of automatic hashtag generation. Please also be aware of some known risks and limitations of our framework. 
To ensure the generalization performance, the S{\scriptsize EG}T{\scriptsize RM} is an end-to-end method relying on large-scale real-world data. 
Such large-scale datasets make the hashtag classification not applicable since it is challenging to unify the generalized hashtag labels. 
Besides, most of the text generation methods generate one hashtag from one input text, and our S{\scriptsize EG}T{\scriptsize RM} is no exception. Thus, S{\scriptsize EG}T{\scriptsize RM} cannot generate an arbitrary number of hashtags for a given input. We would encourage researchers to explore further research and applications of our framework. 
To mitigate the risks mentioned earlier and limitations and improve the real-world usability, we also welcome all kinds of improvements and enhancements from any research field with using our framework.

\section*{Acknowledgment}

This work is supported in part by the NSFC through grant No. U20B2053.


\normalem

\printcredits
\bibliographystyle{unsrtnat}


\bibliography{cas-refs}

\end{document}